\documentclass[runningheads]{llncs}
\pdfoutput=1

\usepackage{graphicx}
\usepackage{siunitx}
\usepackage{booktabs}
\usepackage{dsfont}
\usepackage[inline]{enumitem}
\usepackage{pifont}
\usepackage{xcolor, colortbl}
\usepackage[normalem]{ulem}
\newcommand{\cmark}{\ding{51}}
\newcommand{\xmark}{\ding{55}}
\definecolor{Gray}{gray}{0.9}
\definecolor{LightCyan}{rgb}{0.88,1,1}
\usepackage{tikz}
\usepackage{comment}
\usepackage[caption=false]{subfig}
\usepackage{amsmath,amssymb} 
\usepackage{bm}
\usepackage{makecell}
\usepackage{color}
\usepackage{orcidlink}

\usepackage[accsupp]{axessibility}  % Improves PDF readability for those with disabilities.

\usepackage[capitalize]{cleveref}
\crefname{section}{Sec.}{Secs.}
\Crefname{section}{Section}{Sections}
\Crefname{table}{Table}{Tables}
\crefname{table}{Tab.}{Tabs.}

\renewcommand{\S}{\mathcal{I}}
\newcommand{\R}{\mathcal{R}}
\newcommand{\Z}{\mathcal{Z}}

\newcommand{\xs}{\bm{x}^{\S}_i}
\newcommand{\xr}{\bm{x}^{\R}_i}
\newcommand{\ys}[1][k]{\bm{y}^{\S}_{i,#1}}
\newcommand{\yr}[1][m]{\bm{y}^{\R}_{i,#1}}
\newcommand{\ygs}{\bar{\bm{y}}^{\S}_{i}}
\newcommand{\ygr}{\bar{\bm{y}}^{\R}_{i}}
\newcommand{\zs}[1][k]{\bm{z}^{\S}_{i,#1}}
\newcommand{\zr}[1][m]{\bm{z}^{\R}_{i,#1}}

\newcommand{\zgs}[1][i]{\bar{\bm{z}}^{\S}_{#1}}
\newcommand{\zgr}[1][i]{\bar{\bm{z}}^{\R}_{#1}}
\newcommand{\zsr}[1][m]{\bm{z}^{\S \rightarrow \R}_{i,#1}}
\newcommand{\zrs}[1][k]{\bm{z}^{\R \rightarrow \S}_{i,#1}}

\newcommand{\Es}{E^{\S}}
\newcommand{\Er}{E^{\R}}
\newcommand{\fs}{f^{\S}}
\newcommand{\fr}{f^{\R}}

\newcommand{\fgs}{\bar{f}^{\S}}
\newcommand{\fgr}{\bar{f}^{\R}}
\newcommand{\Asr}{A^{\S \rightarrow \R}}
\newcommand{\Ars}{A^{\R \rightarrow \S}}

\newcommand{\weights}[1][k]{w^{\S}_{i, #1}}
\newcommand{\weightr}[1][m]{w^{\R}_{i, #1}}
\newcommand{\alphars}{\alpha^{\R \rightarrow \S}_{i, k, m}}
\newcommand{\alphasr}{\alpha^{\S \rightarrow \R}_{i, m, k}}
\newcommand{\ps}{p^{\S}_{k, l}}

\newcommand{\ds}{d^{\S}}
\newcommand{\dr}{d^{\R}}
\newcommand{\dz}{d^{\Z}}

\newcommand{\dgz}{\bar{d}^{\Z}}

\makeatletter
\newcommand\footnoteref[1]{\protected@xdef\@thefnmark{\ref{#1}}\@footnotemark}
\makeatother

\begin{document}
\pagestyle{headings}
\mainmatter

\title{Joint Learning of Localized Representations from Medical Images and Reports}
\titlerunning{Localized Representations from Medical Images and Reports}
\author{Philip M\"uller\inst{1}\orcidlink{0000-0001-8186-6479} \and
Georgios Kaissis\inst{1,2,3}\orcidlink{0000-0001-8382-8062} \and
Congyu Zou\inst{4} \and
Daniel Rueckert\inst{1,3}\orcidlink{0000-0002-5683-5889}
}
\authorrunning{P. Müller et al.}
\institute{Institute of Artificial Intelligence in Medicine, Technical University of Munich, 81675 Munich, Germany \\
\email{philip.j.mueller@tum.de}
\and
Institute of Radiology, Technical University of Munich, 81675 Munich, Germany \and
Department of Computing, Imperial College London, London SW7 2BX, UK \and
Department for Internal Medicine I, Klinikum Rechts der Isar, Technical University of Munich, 81675 Munich, Germany}
%\end{comment}
%******************

\maketitle

%%%%%%%%% ABSTRACT
\begin{abstract}
Contrastive learning has proven effective for pre-training image models on unlabeled data with promising results for tasks such as medical image classification. Using paired text (like radiological reports) during pre-training improves the results even further. 
Still, most existing methods target image classification
downstream tasks and may not be optimal for localized tasks like semantic segmentation or object detection. We therefore propose \emph{\textbf{Lo}calized representation learning from \textbf{V}ision and \textbf{T}ext (LoVT)}, to our best knowledge, the first text-supervised pre-training method that targets localized medical imaging tasks. Our method combines instance-level image-report
contrastive learning with local contrastive learning on image region and report sentence representations. We evaluate LoVT and commonly used
pre-training methods on an evaluation framework of 18 localized tasks on chest X-rays from five public datasets. LoVT performs best
on 10 of the 18 studied tasks making it the preferred method of choice
for localized tasks.
\end{abstract}

%%%%%%%%% BODY TEXT
\section{Introduction and Motivation}
\label{sec:intro}
In medical applications of computer vision, high-quality annotated data is scarce and expensive to acquire, as manually labeling samples typically requires trained
physicians\cite{deep_medicine_challenges}. Therefore, the requirement for large labeled datasets can become quite problematic and may limit the applications of deep learning in this field.
One approach to overcome this problem is to utilize radiological reports that are paired with medical images.
Such reports are produced routinely in clinical practice and are typically written by medical experts (e.g. radiologists). They thus provide a valuable source of semantic information that is available with little additional costs.
Rule-based Natural Language Processing (NLP) models like CheXpert\cite{chexpert} extract labels from these reports allowing the automatic creation of large datasets but they also have some significant limitations. Most importantly, such approaches are typically limited to classification tasks. They generate overall labels for reports (and therefore the paired images) but relating these labels to specific image regions is nontrivial so they cannot be used for localized tasks like semantic segmentation or object detection. 
Also, rule-based NLP models have to be manually created and cannot generalize to different classification tasks or even different report writing styles\cite{chexpert}.
Instead of using these reports to generate classification labels, the reports can be utilized directly in the pre-training method, as was first proposed in the ConVIRT method\cite{ConVIRT}. 
Here, the semantic information contained in the reports is used as weak supervision to pre-train image models that are then fine-tuned on labeled downstream tasks, where results can be improved or the number of labeled samples can be reduced. 
We argue that while this approach is quite promising it is not designed for localized downstream tasks. For example, ConVIRT\cite{ConVIRT} only works on per-sample image representations and does not explicitly provide more localized representations that might be beneficial for localized tasks like semantic segmentation and object detection. 
In this work, we therefore study how pre-training methods perform on localized tasks and develop a novel pre-training method designed for localized tasks.

Our contributions are as follows:
\begin{itemize}
    \item We propose a local contrastive loss allowing to align local representations of sentences or image regions while encouraging spatial smoothness and sensitivity.
    \item We split each report into sentences and each image into regions (i.e. patches), compute representations for sentences and regions and align them using an attention mechanism and our proposed local contrastive loss.
    \item We compute global (i.e. per-image and per-report) representations using attention-pooling on the region and sentence representations, and then use a global contrastive loss to align them.
    \item We propose \emph{\textbf{Lo}calized representation learning from \textbf{V}ision and \textbf{T}ext (LoVT)},
a pre-training method that extends ConVIRT\cite{ConVIRT} using our proposed ideas and outperforms it on most localized downstream tasks.
    \item We evaluate our method trained using MIMIC-CXR\cite{MIMIC-CXR-2,MIMIC-CXR,MIMIC-CXR-JPG,PhysioNet} on a downstream evaluation framework\cite{eval} with 18 localized tasks on chest X-rays, including object detection and semantic segmentation on five public datasets. We compare it with several self- and text-supervised methods and with transfer from classification in more than 1400 evaluation runs. Our method LoVT proves as the most successful method outperforming all other methods on 10 out of 18 tasks.
\end{itemize}

\section{Related Work}
In recent years, contrastive learning\cite{nonparam_disicrim,CPC,CPC2,DIM,misra2019selfsupervised,li2021prototypical,SimCLR,MoCo,BYOL,SimSiam,BarlowTwins,VICReg,W-MSE,DINO,SwAV}, has become the state-of-the-art approach for self-supervised representation learning on images. It has been successfully applied as pre-training method in medical imaging including downstream tasks such as image classification on chest X-rays\cite{gazda2021selfsupervised,sowrirajan2021mococxr,sriram2021covid19}.

Most contrastive learning approaches use, unlike our method, only instance-level contrast, i.e.\ represent each view of the image by a single vector. While the resulting representations are well-suited for global downstream tasks, they are not designed for localized downstream tasks. Therefore, there is a number of recent approaches that use region-level contrast\cite{PixelPro,detco,DenseCL,chaitanya2020contrastive,pinheiro2020unsupervised,crosspixel_opticalflow}, i.e.\ they act on representations of image regions. Unlike our method, these methods do not utilize paired text.

Recently however, there is much focus on self-supervised representation learning methods that pre-train image models for downstream tasks by taking advantage of the companion text\cite{CLIP,ALIGN,ConVIRT,VirTex,ICMLM,liu2021loctex}.
VirTex\cite{VirTex} and ICMLM\cite{ICMLM} use image captioning tasks (generative tasks). ConVIRT\cite{ConVIRT}, CLIP\cite{CLIP} and ALIGN\cite{ALIGN} on the other hand use multiview contrastive learning\cite{AMDIM}. These approaches have been found to be more effective for discriminative downstream tasks\cite{CLIP}.
ConVIRT, CLIP, and ALIGN all follow the same general framework where an image and a text encoder are trained jointly using the NT-Xent loss (which is also used in SimCLR) on image and text views. The text views are based on single sentences from companion text, in the case of ConVIRT it is a sentence sampled from the radiology report. The main difference between these methods is the datasets they are studied on, ConVIRT is trained on chest X-rays while the other methods use natural images. Additionally, CLIP uses attention pooling to compute image representations from feature maps while the other methods use the default pooling method from the image encoder (average pooling in the case of ResNet50\cite{ResNet50}).
Our method follows a similar framework but adds local contrastive losses for better performance on localized tasks. Also, it encodes the whole report instead of sampling a single sentence and uses attention pooling in the image and text encoders.
LocTex\cite{liu2021loctex} does localized pre-training on natural images with companion text and predicts alignment of text and image regions. Unlike our method, it uses supervision generated by mouse gazes instead of learning the alignment implicitly using a local contrastive loss.
Most related to our work is the recently published local Mutual Information approach \cite{local_MI} that performs contrastive learning on report sentences and image regions but targets classification instead of localized tasks and does therefore neither encourage contrast between regions nor spatial smoothness.

\section{Method}
\begin{figure}[t]
  \centering
  \scriptsize
  %auto-ignore
\begin{tabular}{ll}
    \toprule
    \textbf{EXAMINATION:} & CHEST (PA and LAT) \\ %CHEST (PORTABLE AP) \\
    \midrule
    \textbf{INDICATION:} & \_\_\_ year old woman with ?pleural effusion \\ %\_\_\_F with cough  // acute process? \\
    %\midrule
    %\textbf{TECHNIQUE:} & Chest PA and lateral \\
    %\textbf{COMPARISON:} & Chest radiograph \_\_\_ \\
    \midrule
    \textbf{FINDINGS:} \\
    \multicolumn{2}{p{.5\linewidth}}{Cardiac size cannot be evaluated.} \\
    \multicolumn{2}{p{.5\linewidth}}{Large left pleural effusion is new.} \\
    \multicolumn{2}{p{.5\linewidth}}{Small right effusion is new.} \\
    \multicolumn{2}{p{.5\linewidth}}{The upper lungs are clear.} \\
    \multicolumn{2}{p{.5\linewidth}}{Right lower lobe opacities are better seen in prior CT.} \\
    \multicolumn{2}{p{.5\linewidth}}{There is no pneumothorax.} \\
    \multicolumn{2}{p{.5\linewidth}}{There are mild degenerative changes in the thoracic spine.} \\
    %\multicolumn{2}{p{.8\linewidth}}{Single frontal view of the chest provided.} \\
    %\multicolumn{2}{p{.8\linewidth}}{There is no focal consolidation, effusion, or pneumothorax.} \\
    %\multicolumn{2}{p{.8\linewidth}}{The cardiomediastinal silhouette is normal.} \\
    %\multicolumn{2}{p{.8\linewidth}}{Again seen are multiple clips projecting over the left breast and remote left-sided rib fractures.} \\ 
    %\multicolumn{2}{p{.8\linewidth}}{No free air below the right hemidiaphragm is seen.} \\
    \midrule
    \textbf{IMPRESSION:} \\
    \multicolumn{2}{p{.5\linewidth}}{Large left pleural effusion.} \\
    %\multicolumn{2}{p{.8\linewidth}}{No acute intrathoracic process.} \\
    \bottomrule
  \end{tabular}
  \caption{Example radiology report describing chest X-Rays. Taken from the MIMIC-CXR\cite{MIMIC-CXR-2,MIMIC-CXR,PhysioNet} dataset.}
  \label{fig:report_example}
\end{figure}
\begin{figure*}[t]
    \centering
    \includegraphics[width=.6\linewidth]{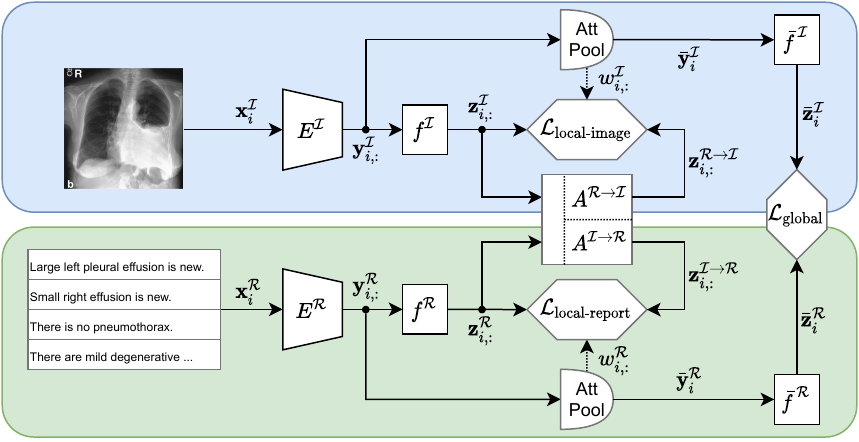}
        \caption{Architecture of LoVT. Given an image $\xs$ and the related report $\xr$, the encoders $\Es$ and $\Er$ compute image region and report sentence representations, respectively, which are projected using $\fs$ and $\fr$. The alignment models $\Ars$ and $\Asr$ compute cross-modal report-to-image ($\zrs$) and image-to-report ($\zsr$) representations which are aligned with the uni-modal representations ($\zs$ and $\zr$) using the local losses $\mathcal{L}_\text{local-image}$ and $\mathcal{L}_\text{local-report}$, respectively. Global image ($\ygs$) and report ($\ygr$) representations are computed using attention pooling on the local representations, are then projected using $\fgs$ and $\fgr$ and aligned using the global loss $\mathcal{L}_\text{global}$.}
        \label{fig:architecture}
\end{figure*}
\subsection{Assumptions and Intuition}\label{sec:assumptions}
As shown in \cref{fig:report_example}, a radiology report is typically split into several sections, including a \emph{Findings} section, describing related radiological images, and an \emph{Assessment} section, interpreting the findings. 
As these sections describe medical aspects observed ({\em Findings}) in one or more related images and conclusions ({\em Assessment}) drawn from it, they provide supervision for identifying relevant patterns in the images and interpretations of these patterns.
Both sections can be split into sentences and each of these sentences typically describes one or a few aspects of which we assume that most are related to one or a few very localized regions in a paired image. We randomly sample one of the images related to a given report and split it into $7 \times 7$ equally-sized regions. More precisely, we augment and resize the image to a size of $224 \times 224$, feed it into a convolutional neural network, and use the output feature map of size $7 \times 7$ as region representations. A language model encodes the tokens of the report as contextualized (i.e.\ considering their meaning in the whole report) vector representations from which we compute sentence representations.
A many-to-many alignment model is then used to compute \emph{cross-modal representations} from \emph{uni-modal representations}, i.e.\ image region representations from sentence representations and vice-versa. 
We argue that by aligning cross-modal and uni-modal representations, the image region representations are encouraged to contain the high-level semantics present in the report.

\subsection{Model Overview}
\cref{fig:architecture} shows the general architecture of our proposed LoVT model.
Each training sample $\bm{x}_i$ is a pair of an image $\xs \in \mathbb{R}^{224 \times 224}$ and the related report $\xr$ consisting of $M_i$ sentences.
Both, $\xs$ and $\xr$, are encoded independently into two global representations, for image and report respectively, and multiple local representations per sample, corresponding to image regions and report sentences, respectively.
An attention-based alignment model then computes cross-modal representations (i.e.\ sentence representations from image regions and vice-versa) which are aligned with the local uni-modal representations using local contrastive losses. Additionally, the global representations are aligned using a global contrastive loss.
The encoders and the alignment model are trained jointly on batches of image-report pairs $\bm{x}_i$.
The details of the model and the loss function will be described in the following sections.

\subsection{Encoding}
Each image $\xs$ is encoded into $K = H \times W$ (we use $K = 7 \times 7$) region representations $\ys \in \mathbb{R}^{\ds}$ using the image encoder $\Es$,
where $k$ is the index of the image region, and $\ds$ is the dimension of the image region representation space.
Our approach is encoder agnostic, i.e.\ any model encoding image regions into vector representations can be used for $\Es$. We use a ResNet50\cite{ResNet50} and take the feature map before global average pooling as region representations.
Similarly, each report $\xr$ is encoded into $M_i$ sentence representations $\yr \in \mathbb{R}^{\dr}$ using the report encoder $\Er$.
Here $M_i$ is the number of sentences of report sample $i$, $m$ is the index of the sentence, and $\dr$ is the dimension of the report sentence representation space. Note that while $K$ is constant, $M_i$ may be different for each sample.
Any model encoding sentences into vector representations can be used for $\Er$. We use BERT\_base\cite{BERT} to jointly encode the tokens of the concatenated sentences of each report and then perform max pooling over the token representations of each sentence to get sentence representations.

The global (i.e.\ per-sample) representations $\ygs$ and $\ygr$ are each computed by an attention pooling layer (not shared between modalities) on the region and sentence representations, respectively. It is implemented using multi-head query-key-value attention\cite{transformer} where the query is computed from the globally averaged region or sentence representations.
This pooling approach was first proposed for the image encoder of CLIP\cite{CLIP}.

Following previous works\cite{ConVIRT,SimCLR,BYOL}, we compute projected local representations $\zs \in \mathbb{R}^{\dz}$ and $\zr \in \mathbb{R}^{\dz}$, and projected global representations $\zgs \in \mathbb{R}^{\dgz}$ and $\zgr \in \mathbb{R}^{\dgz}$ from the representations $\ys$, $\yr$, $\ygs$, and $\ygr$, using the (non-shared) nonlinear transformations $\fs$, $\fr$, $\fgs$, and $\fgr$, respectively,
where $\dz$ is the dimension of the shared local and $\dgz$ of the shared global representation space (we use 512 for both). Note that for local representations the projections are applied to each region $k$ or sentence $m$ independently.

\subsection{Alignment Model}
Following our assumptions (see \cref{sec:assumptions}), we compute an alignment of image regions and sentences and compute cross-modal representations using the alignment models $\Asr$ and $\Ars$, which are based on single-head query-key-value attention\cite{transformer}.

For each sentence $m$ the cross-modal representation $\zsr$ is computed by letting $\zr$ attend to all image region representations $\zs$ (of the related image).
We therefore compute the probability $\alphasr$ that sentence $m$ is aligned with region $k$ based on the scaled dot product scores of their projected representations, i.e.\ $\alphasr = \text{softmax}_{k}\left(\frac{(\bm{Q}\zr)^T(\bm{Q}\zs)}{\sqrt{\dz}}\right)$, where the linear query-key projection $\bm{Q}$ is a learned matrix.
Then the alignment model $\Asr$ uses $\alphasr$ to compute $\zsr$ as projected weighted sum of the image region representations $\zs$:
\begin{align}
\begin{split}
    \zsr = \bm{O} \left(\sum_{k=1}^{K} \alphasr \left(\bm{V}\zs\right)\right)\,,
\end{split}
\end{align}
where the value projection $\bm{V}$, and the output projection $\bm{O}$ are learned matrices.

In a similar fashion the cross-modal representations $\zrs$ are computed by $\Ars$:
\begin{align}
\begin{split}
    \zrs = \bm{O} \left(\sum_{m=1}^{M_i} \alphars \left(\bm{V}\zr\right)\right)\,,
\end{split}
\end{align}
with $\alphars = \text{softmax}_{m}\left(\frac{(\bm{Q}\zs)^T(\bm{Q}\zr)}{\sqrt{\dz}}\right)$. Note that as $\Ars$ and $\Asr$ share the same matrices $\bm{Q}$, $\bm{V}$, and $\bm{O}$, the only difference between $\alphars$ and $\alphasr$ is transposition and the index over which softmax is applied.

\subsection{Loss Function}\label{sec:loss}
\paragraph{Global Alignment}
For global alignment we follow ConVIRT\cite{ConVIRT} and maximize the cosine similarity between paired image and report representations while minimizing the similarity between non-paired (i.e.\ from different samples) representations. The loss consists of a image-report part, where all non-paired report representations from the batch are used as negatives:
\begin{align}
    \ell^{\S\|\R}_\text{global} &=- \log \frac{
        e^{\cos\left(\zgs, \zgr\right) / \tau}
    }{
        \sum_{j} e^{\cos\left(\zgs, \zgr[j]\right) / \tau}}\,,
\end{align}
and a report-image part, defined analogously:
\begin{align}
    \ell^{\R\|\S}_\text{global} &=- \log \frac{
        e^{\cos\left(\zgr, \zgs\right) / \tau}
    }{
        \sum_{j} e^{\cos\left(\zgr, \zgs[j]\right) / \tau}}\,,
\end{align}
where $\tau$ is the similarity temperature (we use $0.1$) and all logarithms are natural.
Both parts are combined using the hyperparameter $\lambda \in [0, 1]$ (we use $0.75$):
\begin{align}
    \mathcal{L}_\text{global} &= \frac{1}{N}\sum_{i=1}^N \left[\lambda\cdot \ell^{\S\|\R}_\text{global} + (1 - \lambda)\cdot \ell^{\R\|\S}_\text{global}\right] \,.
\end{align}

\paragraph{Local Alignment}
The global alignment loss does not only align the global representations but it also prevents the global representations from collapsing to a constant vector using negative samples to contrast the positive pairs. Similarly, we propose local alignment losses encouraging spatial (sentence) sensitivity through negatives from the same sample, i.e.\ preventing the local representations to be similar for all regions (sentences) of an image (report). We use two NT-Xent-based\cite{SimCLR} local losses: $\mathcal{L}_\text{local-image}$, aligning region representations $\zs$ with $\zrs$, and $\mathcal{L}_\text{local-report}$, aligning sentence representations $\zr$ with $\zsr$. 

Some regions or sentences may not be relevant for aligning a sample (e.g.\ background regions or sentences not related to the image). Therefore, we introduce region weights $\weights$ and sentence weights $\weightr$, which are computed as the attention probabilities from the respective attention pooling layer (which was used to compute global representations), averaged over all attention heads. These weights are used in the local loss functions such that irrelevant representations do not have to be aligned. Note that we do not backpropagate through the region or sentence weights.

The loss $\mathcal{L}_\text{local-image}$ allows for having multiple positive pairs within each sample by giving each pair of regions $(k, l)$ a positiveness probability $\ps \in [0, 1]$. 
We then treat each positive pair as its own (weighted) example and contrast it with all other pairs (again all logarithms are natural):
\begin{align}
    \ell^{\S\|\R\rightarrow \S}_\text{local-image} = -\sum_{l=1}^K \ps \log \frac{
        e^{\cos\left(\zs, \zrs[l]\right) / \tau'}
    }{
        \sum_{k'} e^{\cos\left(\zs, \zrs[k']\right) / \tau'}}
\end{align}
\begin{align}
    \ell^{\R\rightarrow \S\|\S}_\text{local-image} = -\sum_{l=1}^K \ps \log \frac{
        e^{\cos\left(\zrs, \zs[l]\right) / \tau'}
    }{
        \sum_{k'} e^{\cos\left(\zrs,  \zs[k']\right) / \tau'}}
\end{align}
\begin{align}
    \mathcal{L}_\text{local-image} &= \frac{1}{2N}\sum_{i=1}^N \sum_{k = 1}^{K} \weights\cdot\left[\ell^{\S\|\R\rightarrow \S}_\text{local-image} + \ell^{\R\rightarrow \S\|\S}_\text{local-image}\right]\,.
\end{align}
Here $\tau'$ is the similarity temperature and is set to $0.3$.
We assume that nearby image regions are often similar and that therefore nearby regions are more likely to be positives while distant regions are more likely to be negatives.
Thus, we define the positiveness probability $\ps$ of two image regions as the complementary cumulative exponential distribution of $d_{\bm{x}}$ (their spatial $\ell_2$-distance in 2D space normalized by the length of the diagonal $\sqrt{H^2 + W^2}$) and set $\ps$ to zero above cutoff threshold $T \in [0, \infty)$:
\begin{align}
    \ps = 
    \frac{
        \mathds{1}_{[d_{\bm{x}}(k, l) \leq T]} \cdot e^{-d_{\bm{x}}(k, l) / \beta}
        }{
        \sum_{k'} \mathds{1}_{[d_{\bm{x}}(k, k') \leq T]} \cdot e^{-d_{\bm{x}}(k, k') / \beta}
    }\,.
\end{align}
Here $\beta \in (0, \infty)$ is a sharpness hyperparameter. We set $\beta=1$ and $T = 0.5$. Note that the normalization of $d_{\bm{x}}$ is equal to rescaling $T$ and $\beta$, i.e.\ it allows us to define both hyperparameters independently of the image size.

The definition of $\ps$ is derived by modeling the occurrence of related features at specific distances in the image as a Poisson point process, such that the $\ell_2$-distance of related features follows the exponential distribution. We assume a Poisson process due to its property of being memoryless, i.e.\ knowing that a feature is already related to another feature at some distance does not change how distant additional related features can be found.
Also, the probability density function of the exponential distribution is decreasing (with support on the interval $[0, \infty)$), which seems reasonable as it is typically more likely that related features are near than far. Its cumulative distribution function then describes the probability that two related features are within a given radius and its complementary function that of being outside a given radius. The threshold $T$ assures that very distant pairs do not count as positives. 
The loss $\mathcal{L}_\text{local-image}$ thus encourages spatial smoothness of image regions while maintaining spatial sensitivity through negative samples.
Note that it is related to the pixel-contrast loss proposed in\cite{PixelPro}, where the main novelty of our work is the partly smooth definition of $\ps$ based on the exponential distribution.

The local report loss $\mathcal{L}_\text{local-report}$ is defined similarly but we do not assume prior knowledge about the similarity of sentences and therefore only have a single positive pair per sentence (again all logarithms are natural):
\begin{align}
    \ell^{\R\|\S\rightarrow \R}_\text{local-report} = -\log \frac{
        e^{\cos\left(\zr, \zsr[m]\right) / \tau'}
    }{
        \sum_{m'} e^{\cos\left(\zr, \zsr[m']\right) / \tau'}}
\end{align}
\begin{align}
    \ell^{\S\rightarrow \R\|\R}_\text{local-report} = -\log \frac{
        e^{\cos\left(\zsr, \zr[m]\right) / \tau'}
    }{
        \sum_{m'} e^{\cos\left(\zsr,  \zr[m']\right) / \tau'}}
\end{align}
\begin{align}
    \mathcal{L}_\text{local-report} &= \frac{1}{2N}\sum_{i=1}^N \sum_{m = 1}^{M_i} \weightr\cdot\left[\ell^{\R\|\S\rightarrow \R}_\text{local-report} + \ell^{\S\rightarrow \R\|\R}_\text{local-report}\right]\,
\end{align}

\paragraph{Total Loss}
The total loss $\mathcal{L}$ is computed as the weighted sum of global and local losses:
\begin{align}
    \mathcal{L} &= \gamma \cdot \mathcal{L}_\text{global} + \mu  \cdot \mathcal{L}_\text{local-image} + \nu  \cdot \mathcal{L}_\text{local-report}\,,
\end{align}
where $\gamma$, $\mu$, and $\nu$ are loss weights to balance the individual losses and are set to $1.0$, $0.75$, and $0.75$, respectively. We determined these loss weights by running small grid searches (see \cref{sec:pre_training_details} for details).

\section{Evaluation}
\subsection{Downstream Tasks and Experimental Setup}
\label{sec:downstream_tasks}
We evaluate our method on a downstream evaluation framework\cite{eval} with 18 localized tasks on chest X-rays, which we will shortly describe here. For more details, we refer to \cref{sec:downstream_eval_details}. 

\paragraph{Evaluation Protocols}
We only use the pre-trained ResNet50 (from the image encoder).
For semantic segmentation tasks we evaluate in the
following settings::
\begin{enumerate*}[label=(\roman*)]
    \item \textbf{U-Net Finetune}: Here the ResNet50 is used as backbone of a U-Net\cite{UNet} and is finetuned jointly with all other layers,
    \item \textbf{U-Net Frozen}: Here the ResNet50 is used as frozen backbone of a U-Net\cite{UNet} and only the non-backbone layers are finetuned, and
    \item \textbf{Linear}: Here an element-wise linear layer is trained that is applied after the last feature map (before pooling) of the frozen ResNet50, before the results are upsampled to the segmentation resolution.
\end{enumerate*}

For object detection tasks we use the following protocols: \begin{enumerate*}[label=(\roman*)]
    \item \textbf{YOLOv3 Finetune}: Here the ResNet50 is used as backbone of a YOLOv3\cite{YOLOv3} model and is finetuned jointly with the non-backbone layers, 
    \item \textbf{YOLOv3 Frozen}: Here the ResNet50 is used as frozen backbone of a YOLOv3\cite{YOLOv3} model and only the non-backbone layers are finetuned, and
    \item \textbf{Linear}: Here the object detection ground truth is converted to segmentation masks and then the \emph{Linear} segmentation protocol is used for evaluation.
\end{enumerate*}

\paragraph{Downstream Datasets}
We evaluate the pre-trained ResNet50 on several medical datasets, namely 
\begin{enumerate*}[label=(\roman*)]
    \item \textbf{RSNA Pneumonia Detection}\cite{NIH_CXR,RSNA_2}, with more than $260000$ 
    frontal-view chest X-rays with detection targets for pneumonia opacities. We use the \emph{YOLOv3 Finetune}, \emph{YOLOv3 Frozen}, and \emph{Linear} protocols, each with $1 \%$, $10 \%$, and $100 \%$ of the training samples;
    \item \textbf{COVID Rural}\cite{COVID_Rural_1,COVID_Rural_2}, with more than $200$ 
    frontal-view chest X-rays with segmentation masks for COVID-19 lung opacity regions. We use the \emph{UNet Finetune}, \emph{UNet Frozen}, and \emph{Linear} protocols;
    \item \textbf{SIIM-ACR Pneumothorax Segmentation}\cite{siim_pneumo}, with more than $12000$ 
    frontal-view chest X-rays with segmentation masks for pneumothorax. We use the \emph{UNet Finetune}, \emph{UNet Frozen} protocols, but due not use \emph{Linear} due to the fine-grained nature of the segmentation masks;
    \item \textbf{Object CXR}\cite{object_CXR} with $9000$ 
    frontal-view chest X-rays with detection targets for foreign objects. We use the \emph{YOLOv3 Finetune}, \emph{YOLOv3 Frozen}, and \emph{Linear} protocols;
    \item \textbf{NIH CXR}\cite{NIH_CXR}, with almost $1000$ 
    frontal-view chest X-rays with detection targets for eight pathologies (Atelectasis, Cardiomegaly, Effusion, Infiltrate, Mass, Nodule, Pneumonia, and Pneumothorax). Due to the limited data per class, we only use the \emph{Linear} protocol.
\end{enumerate*}
The different evaluation protocols complement each other: While the \emph{U-Net Finetune} and \emph{YOLOv3 Finetune} protocols evaluate how well the pre-trained image models could be fine-tuned for practical applications, the \emph{Linear} protocols directly evaluate the learned local representations (i.e.\ feature maps) while adding as few parameters as possible and therefore mostly omitting the variance introduced by random initialization during downstream evaluation. The \emph{U-Net Frozen} and \emph{YOLOv3 Frozen} protocols can be seen as middle ground between the two extremes, where representations are frozen but evaluated in a more practical setting (but with many randomly initialized layers). Overall this allows the analysis of many aspects of the pre-trained representations.

\paragraph{Tuning and Evaluation Procedure}
We tune all models on a single downstream task, \emph{RSNA YOLOv3 Frozen 10\%}. Other downstream tasks have not been evaluated during tuning to make sure that models are not biased towards the downstream tasks.
After tuning, each model was evaluated on all downstream tasks. For each task the downstream learning rates were tuned individually per model (using single evaluation runs) before running five evaluations (all using the tuned learning rate). We report the average results of these five runs and their $95\%$-confidence interval. 

\paragraph{Pre-Training Dataset}
We train our method on version 2 of  MIMIC-CXR\cite{MIMIC-CXR-JPG,MIMIC-CXR,MIMIC-CXR-2,PhysioNet} as, to our best knowledge, it is the largest and most commonly used dataset of this kind.
Since all downstream tasks contain only frontal views, we remove all lateral views, such that roughly 21000 training samples remain, each with a report and one or more frontal images.

\paragraph{Baselines}
We compare our method against several baseline methods:
\begin{itemize}
    \item \textbf{Random Init.}: The ResNet50 is initialized using its default random initialization
    \item \textbf{ImageNet\cite{ImageNet} Init.}: The ResNet50 is initialized with weights pre-trained on the ImageNet ILSVRC-2012 task\cite{ImageNet};
    \item \textbf{CheXpert\cite{chexpert}}: The ResNet50 is pre-trained using supervised multi-label binary classification with CheXpert\cite{chexpert} labels on frontal chest X-rays of MIMIC-CXR
    \item \textbf{Global image pre-training methods}: The ResNet50 is pre-trained using the self-supervised pre-training methods SimCLR\cite{SimCLR} or BYOL\cite{BYOL} on frontal chest X-rays of MIMIC-CXR. We decided to include SimCLR as is uses a similar loss function as LoVT and we include BYOL because of its widespread use.
    \item \textbf{Local image pre-training methods}: The ResNet50 is pre-trained using the self-supervised pre-training method PixelPro\cite{PixelPro} on frontal chest X-rays of MIMIC-CXR.
    We include PixelPro to study the effect of local contrastive losses when using only images.
    \item \textbf{Global image-text pre-training methods}:
    The ResNet50 is pre-trained using the image-text methods ConVIRT\cite{ConVIRT} or CLIP\cite{CLIP} on frontal MIMIC-CXR. Note that for comparability we adapted CLIP to use the same image and text encoders as ConVIRT such that the main difference between CLIP and ConVIRT is that CLIP uses attention pooling to compute the scan representation while ConVIRT uses average pooling.
    We include both methods as LoVT builds upon a similar general framework, where we include ConVIRT because it targets chest X-rays (like LoVT) and include CLIP because of its widespread use and as it uses (like LoVT) attention pooling in the image encoder. We decided not to include VirTex\cite{VirTex} and ICMLM\cite{ICMLM} as they use generative tasks, which have been found to be less effective for discriminative downstream tasks\cite{CLIP}.
\end{itemize}

\subsection{Downstream Results}
\label{sec:results}
\begin{table*}[t]
  \centering
  \caption{Results on the RSNA pneumonia detection tasks with different training set sizes. All results are averaged over five evaluation runs and the 95\%-confidence interval is shown. The best results per task are underlined, the second-best results are dash-underlined and the best results per pre-training category (general initialization, pre-training on 30\% and 100\%) are highlighted in bold. Note that the \emph{YOLOv3 Frozen 10\%} task (task 5) was used for tuning of all methods and may therefore not be representative as methods may overfit on this task.}
  \tiny
  \setlength{\tabcolsep}{1.5pt}
  %auto-ignore
\begin{tabular}{lcccc|>{\columncolor{Gray}}c|cccc}
\toprule
& \multicolumn{3}{c}{RSNA YOLOv3 Finetune} & \multicolumn{3}{c}{RSNA YOLOv3 Frozen} & \multicolumn{3}{c}{RSNA Lin.\ Seg.} \\
         & \multicolumn{3}{c}{mAP ($\%$)} & \multicolumn{3}{c}{mAP ($\%$)} & \multicolumn{3}{c}{Dice ($\%$)} \\
                       & 1\% & 10\% & 100\% & 1\% & 10\% & 100\% & 1\% & 10\% & 100\% \\
    \midrule
\multicolumn{10}{l}{\emph{General initialization methods}}\\
Random & 2.4$\pm$0.5 & 5.1$\pm$1.2 & 14.9$\pm$1.7 & 1.0$\pm$0.2 & 4.0$\pm$0.3 & 8.9$\pm$0.9 & 21.9$\pm$1.2 & 5.3$\pm$0.0 & 5.3$\pm$0.0 \\
ImageNet~\cite{ImageNet} & \textbf{5.0$\pm$0.7} & \textbf{12.4$\pm$0.8} & \textbf{19.0$\pm$0.2} & \textbf{3.6$\pm$1.4} & \textbf{8.0$\pm$0.1} & \textbf{15.7$\pm$0.3} & \textbf{27.5$\pm$0.6} & \textbf{38.3$\pm$0.0} & \textbf{43.3$\pm$0.0} \\
\midrule
\multicolumn{10}{l}{\emph{Pre-Training on 30 \% of frontal MIMIC-CXR}}\\
CheXpert~\cite{chexpert} & \textbf{8.3$\pm$0.8} & 12.4$\pm$1.6 & \dashuline{\textbf{21.3$\pm$0.3}} & 7.0$\pm$1.0 & 14.8$\pm$0.8 & 18.8$\pm$0.4 & 38.9$\pm$0.2 & 45.5$\pm$0.2 & 48.1$\pm$0.0 \\
%\_BYOL~\cite{BYOL} & 7.5$\pm$1.2 & 11.7$\pm$1.8 & 17.9$\pm$0.1 & 6.3$\pm$1.2 & 13.9$\pm$1.0 & \underline{\textbf{20.5$\pm$1.2}} & 43.3$\pm$0.1 & 48.9$\pm$0.1 & 50.9$\pm$0.0 \\
%\_SimCLR~\cite{SimCLR} & 6.4$\pm$0.7 & \underline{\textbf{14.9$\pm$0.6}} & 18.3$\pm$2.3 & 6.2$\pm$0.8 & 12.4$\pm$0.4 & 16.5$\pm$0.1 & 42.5$\pm$0.0 & 45.3$\pm$0.0 & 48.0$\pm$0.0 \\
BYOL~\cite{BYOL} & 7.0$\pm$1.0 & 11.9$\pm$1.1 & 18.8$\pm$0.2 & 9.6$\pm$0.2 & 14.0$\pm$1.2 & \textbf{\underline{21.0$\pm$0.2}} & 42.9$\pm$0.1 & 47.8$\pm$0.2 & 50.0$\pm$0.0 \\
SimCLR~\cite{SimCLR} & 6.7$\pm$0.5 & \textbf{12.9$\pm$0.5} & 20.4$\pm$1.8 & 7.9$\pm$1.0 & 11.9$\pm$0.1 & 19.9$\pm$0.2 & 43.1$\pm$0.0 & 46.0$\pm$0.0 & 48.2$\pm$0.0 \\
PixelPro~\cite{PixelPro} & 4.8$\pm$0.6 & 12.6$\pm$1.2 & 19.8$\pm$0.4 & 3.1$\pm$0.2 & 6.4$\pm$0.5 & 13.4$\pm$0.3 & 25.9$\pm$0.2 & 34.6$\pm$0.0 & 39.8$\pm$0.1 \\
ConVIRT~\cite{ConVIRT} & 7.4$\pm$1.3 & 12.7$\pm$1.5 & 18.3$\pm$0.4 & \textbf{\dashuline{9.8$\pm$0.3}} & 14.8$\pm$1.1 & 18.4$\pm$1.1 & 42.1$\pm$0.1 & 47.1$\pm$0.2 & 50.2$\pm$0.0 \\
CLIP~\cite{CLIP}* & 7.2$\pm$0.8 & 12.8$\pm$1.2 & 19.7$\pm$0.5 & 9.3$\pm$0.4 & 16.1$\pm$1.1 & 19.6$\pm$1.4 & 44.3$\pm$0.1 & 48.8$\pm$0.1 & 50.7$\pm$0.0 \\
\rowcolor{LightCyan}
LoVT (Ours) & 7.7$\pm$1.0 & 11.7$\pm$0.5 & 17.2$\pm$1.3 & 8.6$\pm$1.5 & \textbf{\underline{17.9$\pm$0.4}} & 18.0$\pm$0.1 & \textbf{\dashuline{46.0$\pm$0.0}} & \textbf{\dashuline{49.4$\pm$0.0}} & \textbf{\dashuline{51.5$\pm$0.0}} \\
\midrule
\multicolumn{10}{l}{\emph{Pre-Training on 100 \% of frontal MIMIC-CXR}}\\
CheXpert~\cite{chexpert} & \dashuline{10.0$\pm$1.9} & 12.4$\pm$0.9 & \textbf{\underline{22.2$\pm$0.4}} & 5.8$\pm$0.4 & 11.9$\pm$0.7 & 20.0$\pm$0.2 & 40.0$\pm$0.1 & 44.3$\pm$0.0 & 46.9$\pm$0.0 \\
BYOL~\cite{BYOL} & 5.6$\pm$0.8 & 11.0$\pm$0.2 & 17.3$\pm$1.1 & 6.8$\pm$1.6 & 12.1$\pm$1.1 & 15.9$\pm$0.6 & 41.9$\pm$0.0 & 45.1$\pm$0.0 & 46.8$\pm$0.0 \\
SimCLR~\cite{SimCLR} & 7.1$\pm$0.7 & 12.2$\pm$0.8 & 18.8$\pm$1.0 & 5.4$\pm$0.2 & 13.1$\pm$0.2 & 17.3$\pm$1.6 & 43.0$\pm$0.0 & 45.1$\pm$0.0 & 47.0$\pm$0.0 \\
PixelPro~\cite{PixelPro} & 4.8$\pm$0.3 & 11.0$\pm$1.5 & 17.4$\pm$1.7 & 4.6$\pm$1.6 & 5.4$\pm$1.1 & 12.6$\pm$1.3 & 23.9$\pm$0.4 & 34.8$\pm$0.2 & 40.2$\pm$0.1 \\
ConVIRT~\cite{ConVIRT} & \textbf{\underline{10.7$\pm$1.1}} & \textbf{\underline{13.3$\pm$0.8}} & 18.5$\pm$0.4 & 8.2$\pm$0.9 & 15.6$\pm$1.2 & 17.9$\pm$0.3 & 44.6$\pm$0.1 & 48.5$\pm$0.0 & 50.4$\pm$0.3 \\
CLIP~\cite{CLIP}* & 7.0$\pm$1.5 & 10.7$\pm$1.1 & 19.9$\pm$0.8 & \textbf{\underline{11.9$\pm$0.7}} & 15.0$\pm$1.1 & 18.7$\pm$0.0 & 45.2$\pm$0.0 & 49.3$\pm$0.1 & 51.1$\pm$0.0 \\
\rowcolor{LightCyan}
LoVT (Ours) & 8.5$\pm$0.8 & \dashuline{13.2$\pm$0.6} & 18.1$\pm$3.2 & 9.6$\pm$1.2 & \textbf{\dashuline{16.4$\pm$1.3}} & \textbf{\dashuline{20.5$\pm$1.0}} & \textbf{\underline{46.3$\pm$0.0}} & \textbf{\underline{50.1$\pm$0.0}} & \textbf{\underline{51.8$\pm$0.0}} \\
\midrule
Task Nr. & 1 & 2 & 3 & 4 & 5 & 6 & 7 & 8 & 9 \\
\bottomrule
        \end{tabular}\\
  * Modified to use the same image and text encoders as ConVIRT and LoVT.
  \label{tab:results_1}
\end{table*}
\begin{table*}[t]
  \centering
  \caption{Results on downstream tasks on the COVID Rural, SIIM Pneumothorax, Object CXR, and NIH CXR datasets. All results are averaged over five evaluation runs and the 95\%-confidence interval is shown. The best results per task are underlined, the second-best results are dash-underlined and the best results per pre-training category (general initialization, pre-training on 30\% and 100\%) are highlighted in bold.}
  \tiny
  \setlength{\tabcolsep}{1.5pt}
  %auto-ignore
\begin{tabular}{lccccccccc}
\toprule
& \multicolumn{3}{c}{COVID Rural} & \multicolumn{2}{c}{SIIM-ACR Pneumoth.} & \multicolumn{3}{c}{Object CXR} & NIH CXR \\
    & UNet & UNet & Linear & UNet & UNet & YOLOv3 & YOLOv3 & Linear & Linear \\
    & Finetune & Frozen & & Finetune & Frozen & Finetune & Frozen & & \\
    & Dice ($\%$) & Dice ($\%$) & Dice ($\%$) & Dice ($\%$) & Dice ($\%$) & fROC ($\%$) & fROC ($\%$) & Dice ($\%$) & \makecell{Avg \\ Dice ($\%$)} \\
    \midrule
\multicolumn{10}{l}{\emph{General initialization methods}}\\
Random & 34.0$\pm$1.1 & 32.2$\pm$1.8 & 6.0$\pm$0.0 & 23.2$\pm$1.0 & 23.9$\pm$1.6 & 49.5$\pm$1.2 & 28.4$\pm$1.4 & 6.9$\pm$0.0 & 0.5$\pm$0.4 \\
ImageNet~\cite{ImageNet} & \textbf{43.9$\pm$2.0} & \textbf{41.9$\pm$1.7} & \textbf{32.6$\pm$0.7} & \textbf{38.5$\pm$0.9} & \textbf{36.9$\pm$0.7} & \textbf{62.5$\pm$0.4} & \textbf{52.7$\pm$1.3} & \textbf{37.8$\pm$0.0} & \textbf{2.6$\pm$1.6} \\
\midrule
\multicolumn{10}{l}{\emph{Pre-Training on 30 \% of frontal MIMIC-CXR}}\\
CheXpert~\cite{chexpert} & 43.5$\pm$4.9 & 44.1$\pm$3.2 & 32.1$\pm$2.0 & 38.9$\pm$0.9 & 40.7$\pm$0.7 & 62.2$\pm$0.6 & 46.3$\pm$1.9 & 16.5$\pm$7.7 & 8.7$\pm$0.6 \\
BYOL~\cite{BYOL} & 46.2$\pm$1.6 & \dashuline{47.5$\pm$1.6} & 36.9$\pm$1.7 & 43.1$\pm$0.6 & 42.9$\pm$0.3 & 59.6$\pm$1.0 & 55.7$\pm$1.0 & 32.3$\pm$0.1 & 6.0$\pm$0.1 \\
SimCLR~\cite{SimCLR} & 44.9$\pm$2.9 & 41.4$\pm$3.7 & 33.0$\pm$0.0 & 42.6$\pm$0.4 & 39.2$\pm$0.7 & 61.9$\pm$0.8 & 54.3$\pm$1.0 & 33.2$\pm$0.1 & 13.3$\pm$0.5 \\
%\_BYOL~\cite{BYOL} & 46.5$\pm$3.2 & 46.0$\pm$2.0 & 44.8$\pm$0.1 & 45.4$\pm$0.9 & 44.4$\pm$0.4 & 61.5$\pm$1.0 & 56.0$\pm$0.8 & 34.0$\pm$0.2 & 8.7$\pm$0.7 \\
%\_SimCLR~\cite{SimCLR} & 46.1$\pm$4.4 & 39.1$\pm$0.7 & 39.5$\pm$0.0 & 41.2$\pm$0.5 & 41.2$\pm$0.7 & 59.1$\pm$1.9 & 53.4$\pm$0.5 & 35.1$\pm$0.0 & 11.0$\pm$0.0 \\
PixelPro~\cite{PixelPro} & 47.0$\pm$3.4 & 38.5$\pm$3.9 & 26.6$\pm$0.4 & 39.3$\pm$0.8 & 39.1$\pm$0.3 & \dashuline{\textbf{63.1$\pm$0.7}} & 46.3$\pm$0.2 & 29.9$\pm$0.2 & 1.8$\pm$0.0 \\
ConVIRT~\cite{ConVIRT} & 48.8$\pm$2.2 & 44.2$\pm$3.1 & 45.0$\pm$3.0 & 42.5$\pm$1.0 & 42.5$\pm$0.2 & 62.5$\pm$0.1 & 54.0$\pm$0.7 & 37.7$\pm$0.1 & 11.4$\pm$0.8 \\
CLIP~\cite{CLIP}* & 49.3$\pm$2.0 & 46.5$\pm$2.3 & \dashuline{46.2$\pm$0.3} & 42.8$\pm$1.5 & 42.5$\pm$0.6 & 62.9$\pm$0.8 & 55.5$\pm$2.1 & \textbf{39.0$\pm$0.0} & 12.5$\pm$1.0 \\
\rowcolor{LightCyan}
LoVT (Ours) & \textbf{49.5$\pm$1.3} & \textbf{\underline{49.2$\pm$4.6}} & \textbf{\underline{49.2$\pm$0.2}} & \textbf{43.4$\pm$0.7} & \textbf{43.1$\pm$0.6} & 61.0$\pm$1.3 & \textbf{55.8$\pm$1.1} & 37.6$\pm$0.2 & \textbf{13.4$\pm$0.8} \\
\midrule
\multicolumn{10}{l}{\emph{Pre-Training on 100 \% of frontal MIMIC-CXR}}\\
CheXpert~\cite{chexpert} & 46.2$\pm$1.7 & 45.9$\pm$3.9 & 37.7$\pm$0.4 & 34.2$\pm$0.8 & 37.7$\pm$0.3 & 57.5$\pm$1.1 & 39.8$\pm$2.4 & 19.4$\pm$0.1 & \dashuline{15.2$\pm$0.0} \\
BYOL~\cite{BYOL} & \dashuline{50.7$\pm$2.7} & 42.0$\pm$3.0 & 32.9$\pm$0.0 & 42.6$\pm$0.7 & 40.7$\pm$0.7 & 60.6$\pm$1.1 & 53.1$\pm$0.8 & 21.8$\pm$0.1 & 5.7$\pm$0.0 \\
SimCLR~\cite{SimCLR} & 48.1$\pm$2.5 & 44.1$\pm$2.1 & 35.3$\pm$0.0 & 41.2$\pm$0.8 & 38.7$\pm$0.5 & 61.1$\pm$0.7 & 48.7$\pm$0.5 & 30.0$\pm$0.0 & 11.8$\pm$0.0 \\
PixelPro~\cite{PixelPro} & 42.4$\pm$4.4 & 37.7$\pm$1.0 & 18.9$\pm$6.4 & 39.4$\pm$1.2 & 38.7$\pm$0.6 & \textbf{\underline{65.0$\pm$0.5}} & 46.2$\pm$1.2 & 29.7$\pm$0.1 & 1.8$\pm$0.0 \\
ConVIRT~\cite{ConVIRT} & 47.9$\pm$0.7 & 46.0$\pm$1.1 & 42.7$\pm$2.0 & 39.3$\pm$0.3 & 43.1$\pm$0.3 & 60.6$\pm$1.2 & 52.5$\pm$1.0 & 36.0$\pm$0.0 & \textbf{\underline{18.6$\pm$0.1}} \\
CLIP~\cite{CLIP}* & 48.6$\pm$2.4 & 45.8$\pm$4.1 & 41.7$\pm$0.1 & \dashuline{44.0$\pm$0.7} & \underline{\textbf{45.0$\pm$0.5}} & 62.8$\pm$0.5 & \dashuline{56.9$\pm$1.4} & \dashuline{39.4$\pm$0.0} & 11.4$\pm$0.8 \\
\rowcolor{LightCyan}
LoVT (Ours) & \textbf{\underline{51.2$\pm$2.5}} & \textbf{46.2$\pm$2.4} & \textbf{44.0$\pm$0.8} & \underline{\textbf{44.1$\pm$0.3}} & \dashuline{43.9$\pm$0.7} & 62.1$\pm$0.5 & \textbf{\underline{57.4$\pm$0.5}} & \textbf{\underline{39.9$\pm$0.0}} & 9.4$\pm$0.5 \\
\midrule
Task Nr. & 10 & 11 & 12 & 13 & 14 & 15 & 16 & 17 & 18 \\
\bottomrule
        \end{tabular}\\
  * Modified to use the same image and text encoders as ConVIRT and LoVT.
  \label{tab:results_2}
\end{table*}
\begin{figure}[t]
     \centering
     \subfloat{
         \includegraphics[width=.3\linewidth]{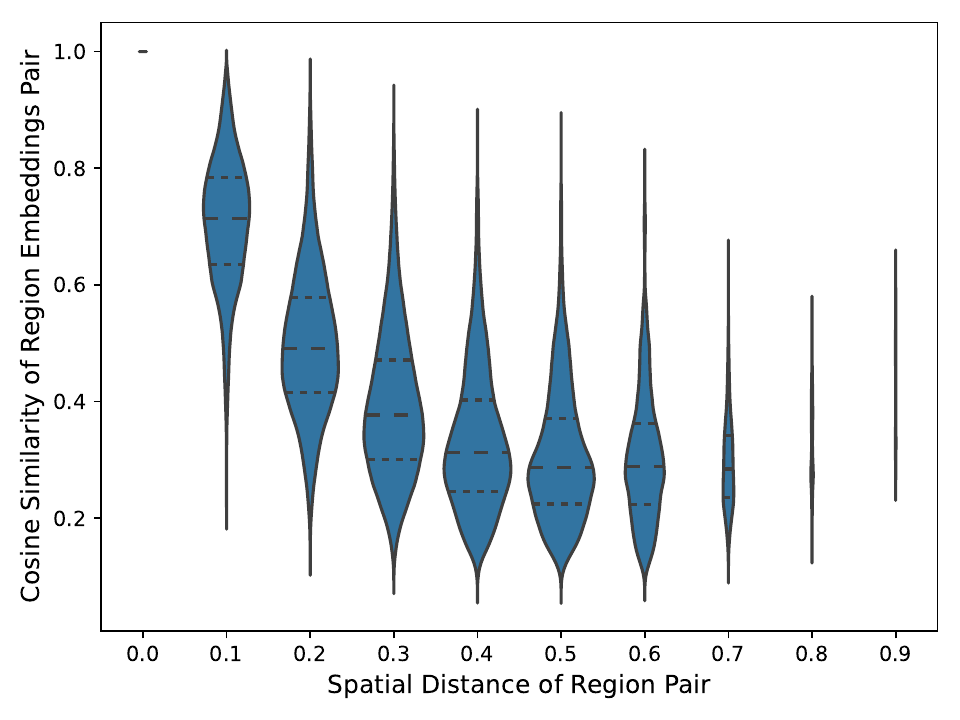}
     }
     \hfill
     \subfloat{
         \includegraphics[width=.3\linewidth]{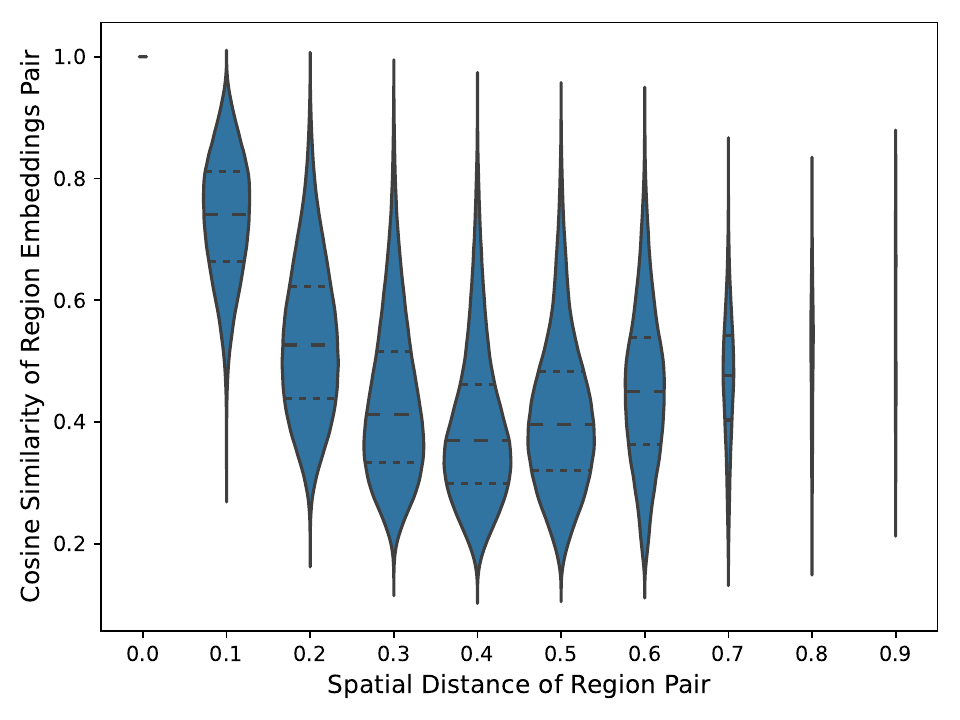}
     }
     \hfill
     \subfloat{
         \includegraphics[width=.3\linewidth]{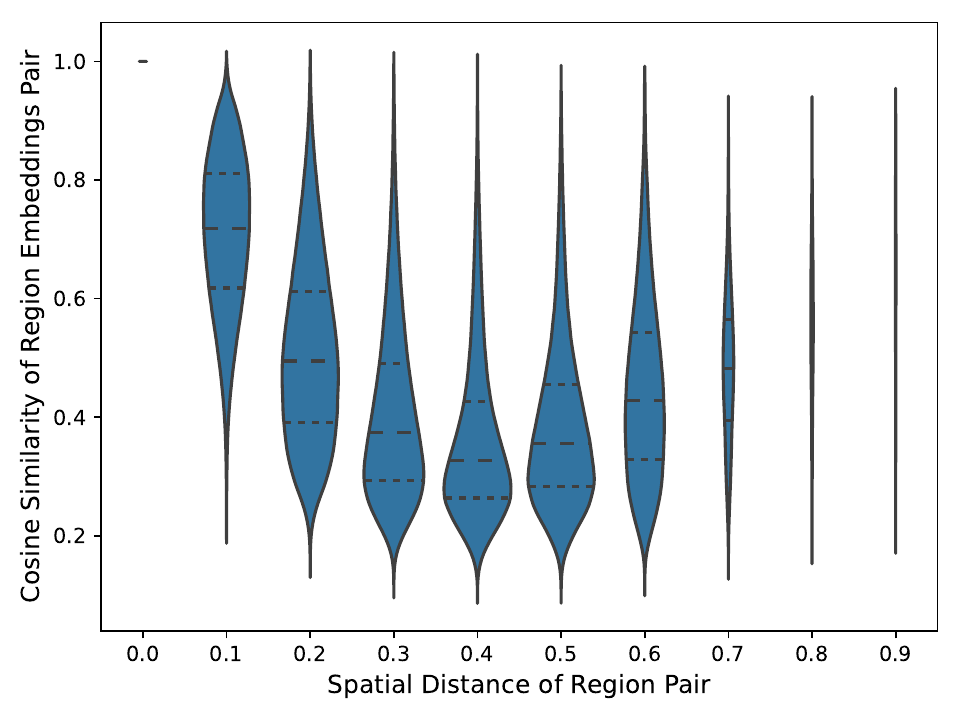}
     }
        \caption{Spatial smoothness and sensitivity of image region representations. \textbf{Left}: LoVT (Ours). \textbf{Middle}: No local losses. \textbf{Right}: No local losses and no attention pooling. Cosine similarities of image region pairs $\ys, \ys[k']$ (each from the same sample) plotted as violin plots (with their width representing the number of pairs and quartiles shown as dashed lines) over their spatial distance in the $7 \times 7$ image space (normalized and rounded to one decimal digit). We trained all models on 30\% of the data and computed the representations on the test set.}
        \label{fig:spatial_smoothness}
\end{figure}

We present the downstream results of our model LoVT and the baselines, with pre-training on 100\% and 30\% of MIMIC-CXR.
\cref{tab:results_1} shows the results on different subsets of the RSNA dataset and \cref{tab:results_2} shows the results on the remaining downstream datasets, i.e.\ on COVID Rural, SIM-ACR Pneumothorax, Object CXR, and NIH CXR.

\paragraph{Comparison of Methods}
We found that there is no single pre-training method performing best on all evaluated downstream tasks. 
On most tasks (15 out of 18) image-text
% task 1, 2, 4, 5, 7, 8, 9, 10, 11, 12, 13, 14, 16, 17, 18
self-supervised methods (i.e.\ LoVT, CLIP, or ConVIRT) outperform the other methods, such that they should be preferred if paired text is available. 

Our model LoVT is the best method (over all pre-training settings) on 10 of 18 tasks,
% tasks 5, 7, 8, 9, 10, 11, 12, 13, 16, 17
%, which is more than any other studied method, 
and significantly outperforms all other methods in 6 of these tasks, % tasks 5, 7, 8, 9, 12, 17
%and within the best two methods on 12 tasks.
% tasks 5, 6, 7, 8, 9, 10, 11, 12, 13, 14, 16, 17,
while the second-best method CLIP significantly outperforms all other methods only on 2 tasks. % tasks 4, 14
LoVT outperforms image-only methods (i.e.\ BYOL, SimCLR, and PixelPro) on 14 tasks, % 1, 2, 5, 7, 8, 9, 10, 11, 12, 13, 14, 16, 17, 18
where the localized image-only method PixelPro outperforms LoVT only on one task (task 15).
On 11 tasks  % 5, 6, 7, 8, 9, 10, 11, 12, 13, 16, 17
LoVT outperforms other text-supervised methods (i.e.\ ConVIRT and CLIP),
on 14 tasks it outperforms CheXpert classification % 2, 4, 5, 6, 7, 8, 9, 10, 11, 12, 13, 14, 16, 17
and on all but two tasks it outperforms ImageNet initialization. % except 3, 15
When using 100\% of the pre-training data LoVT is the best pre-training method on 11 tasks (better by at least the confidence interval on 5 tasks)
% tasks (5), (6), 7, 8, 9, (10), (11), 12, (13), (16), 17 
and when using 30\% on 11 tasks (significantly the best on 4 tasks).
% tasks 5, 7, 8, 9, (10), (11), 12, (13), (14), (16), (18)
LoVT performs best on all COVID Rural tasks, best on most \emph{Linear} tasks, and quite well on the \emph{Frozen} protocol, but does not perform well on the NIH CXR dataset and when finetuned on the RSNA dataset.
As there is no single method performing best on all tasks and LoVT performs best in the majority of tasks, this makes LoVT the default method of choice for localized downstream tasks.

\paragraph{Relevance of Pre-Training Dataset Size}
In our experiments we do not observe a consistent benefit of using roughly 210000 pre-training samples (i.e.\ 100\% of the data) over using roughly 63000 samples (i.e.\ 30\%). While on some datasets like RSNA and Object CXR many methods often perform better when pre-trained on 210000 samples (100\%), on other datasets like COVID Rural, methods often perform better when pre-trained on 63000 samples (30\%). 
When comparing LoVT pre-trained on 30\% of the data 
with other methods pre-trained in both settings (i.e.\ 30\% and 100\%), we observe that LoVT outperforms image-only methods (i.e.\ BYOL, SimCLR, and PixelPro) on 12 tasks, %(1, 5, 7, 8, 9, 11, 12, 13, 14, 16, 17, 18)
other text-supervised methods (i.e.\ ConVIRT and CLIP) on 7 tasks  % 5, 7, 8, 9, 10, 11, 12
and CheXpert classification on 12 tasks,  % 4, 5, 7, 8, 9, 10, 11, 12, 13, 14, 16, 17
showing that LoVT effectively reduces the number of required pre-training samples. 

\paragraph{Relevance of Downstream Dataset Size}
The results shown in \cref{tab:results_1} suggest that, as expected, larger downstream training sets lead to better results. However, we observe that for text-supervised methods (i.e.\ LoVT, CLIP, and ConVIRT), the 
downstream training set size is often less relevant compared to other methods. On the \emph{RSAN YOLOv3 Frozen} tasks, LoVT (100\%) outperforms ImageNet initialization by 31\% when using 100\% of the downstream samples, while it outperforms ImageNet initialization by even 167\% when only using 1\% of the samples. 

\paragraph{Spatial Smoothness and Sensitivity}
We analyze the influence of the local losses and attention pooling on the spatial smoothness and sensitivity of image region representations and therefore plot in \cref{fig:spatial_smoothness} the distributions of the cosine similarity of image region pairs over their spatial distances.
For our LoVT model spatial smoothness and sensitivity can be observed as the quartiles and extreme points of the cosine similarity distributions decrease monotonously with increasing spatial distance, except for a few very distant region pairs with distances larger than $0.6$. Note that these spatially very distant region pairs very likely represent opposite borders (or corners) of the image such that they both very likely contain background, explaining that they have more similar representations. Without local losses $\mathcal{L}_\text{local-image}$ and $\mathcal{L}_\text{local-report}$, the quartiles and extreme points decrease only for small spatial distances while increasing again for points further away, showing that spatial smoothness is only present for nearby regions and spatial sensitivity of more distant region is not optimal. When additionally replacing attention pooling with average (for image regions) and max (for sentences) pooling, similar results can be observed except that the quartiles are decreasing faster and the maximum points do not decrease for nearby regions. We can therefore deduce that the local losses effectively encourage spatial smoothness and sensitivity while attention pooling alone has only little effect.

\paragraph{Analysis of LoVT and Ablation Study}
We refer to \cref{app:analysis} for a detailed analysis of our method LoVT, including an ablation study (focusing on local weighting, global and local losses, and attention pooling), an analysis of the distribution and alignment of learned representations, and an analysis of the region weights $\weights$.

\section{Discussion}
\paragraph{Limitations of our Evaluation Procedure}
We did not tune image encoder, downstream architectures, or preprocessing for downstream tasks, resized all inputs to only $224 \times 224$, and applied no data augmentation. Therefore, the presented downstream results are below results typically reported for these datasets. The evaluation procedure is kept simple to i) limit computational resources, ii) avoid bias induced by downstream tuning, and iii) allow for a fair comparison of pre-training methods, being the main purpose of this work. We assume that benefits observed in our evaluation procedure also indicate benefits for tuned real-world tasks, although they cannot be precisely quantified by our evaluation method.

\paragraph{Limitations of LoVT}
LoVT learns its alignment model implicitly based only on latent representations and instance-level pairing information. This makes the model sensitive to hyperparameters and hard to train. Also, it only uses local negatives from the same sample which restricts the number of negatives and may therefore limit its performance. Additionally, the alignment model is restricted to a simple attention mechanism and the regions are based on fixed patches that are not adaptive to the contents of the image. This may restrict the capabilities of the model and therefore of the pre-training method.
For a detailed discussion of these limitations as well as of the potential negative societal impact we refer to \cref{app:societal_impacts}.

\paragraph{Conclusion}
We study pre-training for localized medical imaging on chest X-rays and propose a novel text-supervised method called LoVT, that combines instance-level contrastive learning with local contrastive learning. 
We evaluate our method on a novel evaluation framework consisting of 18 localized tasks on
chest X-rays and compare it with typically used pre-training and initialization methods. While there is no single best method on all tasks, our method LoVT is the best method on 10 out of 18 studied tasks making it the method of choice for localized tasks.

We hope that our work provides valuable insights that encourage using pre-training for localized medical imaging and that our method inspires future work on localized text-supervised pre-training.

%%%%%%%%% REFERENCES
\bibliographystyle{splncs04}
\bibliography{ms}

\appendix
%auto-ignore
\section{Analysis and Ablation Study}\label{app:analysis}
In this section we analyze the relevance of different components of LoVT (as proposed in the main paper) and study its learned representations.
In order to save computational resources, all further analysis and the ablation study are conducted on $30 \%$ of the pre-training data.

\paragraph{Ablation Study}
\begin{table*}[t]
    \caption{Results of the ablation study evaluated on the \emph{RSNA YOLOv3 Frozen 10\%} task. Results are averaged over five evaluation runs and the $95\%$-confidence interval is shown. The best results are highlighted in bold.}
  \centering
  \tiny
  %auto-ignore
\begin{tabular}{lcccccc}
    \toprule
    Method  & Global loss      & Local losses & \makecell{Local \\ weighting} & Pool      & LR scheduler      & \makecell{RSNA YOLOv3 \\ Frozen 10 \%} \\
    \midrule
    \rowcolor{LightCyan}
    LoVT (Ours)& \cmark           & \cmark     & \cmark        & attention & cyclic-cosine     & \textbf{17.9$\pm$0.4} \\
    \midrule
    %--         & \cmark           & \cmark     & \cmark        & attention & reduce-on-plateau & \\  % no_coslr
    --         & \cmark           & \cmark     & \xmark        & attention & cyclic-cosine     & 16.9$\pm$1.3 \\  % no_weights
    --         & \cmark           & \cmark     & \xmark $^1$      & avg/max   & cyclic-cosine     & 15.7$\pm$0.4 \\  % no_no_att_no_weights
    --  & \xmark           & \cmark     & \xmark $^1$      & --        & cyclic-cosine     & 12.3$\pm$0.7 \\  % no_g
    \midrule
    --         & \cmark           & non-smooth $\mathcal{L}_\text{local-scan}$ & \cmark        & attention & cyclic-cosine     & 15.6$\pm$1.4 \\  % no_spatial
    --         & \cmark           & $\mathcal{L}_\text{local-report}$ only & \cmark        & attention & cyclic-cosine     & 16.2$\pm$0.7 \\  % no_l_scan
    --         & \cmark           & $\mathcal{L}_\text{local-scan}$ only & \cmark        & attention & cyclic-cosine     & 13.7$\pm$1.2 \\  % no_l_report
    --  & \cmark           & \xmark     & --            & attention & cyclic-cosine     & 17.4$\pm$0.9 \\  % no_l
    -- & \cmark           & \xmark     & --            & avg/max   & cyclic-cosine     & 14.2$\pm$1.0 \\  % g_only
    \midrule
    CLIP$^2$         & single sentence  & \xmark     & --            & attention & cyclic-cosine     & 16.1$\pm$1.1 \\  % g_single_sent_att
    --         & single sentence  & \xmark     & --            & avg/max   & cyclic-cosine     & 15.8$\pm$1.3 \\  % convirt_lr
    ConVIRT    & single sentence  & \xmark     &  --           & avg/max   & reduce-on-plateau & 14.8$\pm$1.1 \\
    \bottomrule
  \end{tabular}
  \\ $^1$ Not realizable.
  \\ $^2$ Modified to use the same image and text encoders as ConVIRT and LoVT.
  
  \label{tab:ablations}
\end{table*}

We conduct an ablation study to analyze the relevance of components of LoVT and the effects of the changes we made compared to ConVIRT. Focus of our ablation study are 
\begin{enumerate*}[label=(\roman*)]
    \item the local weighting,
    \item the global and local losses,
    \item and attention pooling.
\end{enumerate*}
We compare different model variants and their results on the \emph{RSNA YOLOv3 Frozen 10\%} task in \cref{tab:ablations}. Note that we focus our ablation study on this single task as this is the task used for tuning all models and baselines while the other tasks are only used for the final evaluation (see \cref{sec:downstream_eval_details}).

Starting from the unmodified LoVT model we first remove the local (region and sentence) weighting in the local losses, i.e.\ we use constant weights $\weights$ and $\weightr$, and observe inferior results, showing the relevance of these weights. 
We then also remove attention pooling and replace it by average (avg) pooling for images and max pooling for reports. The performance further decreases highlighting the importance of attention pooling. Note that the local weights cannot be computed without attention pooling, making a model with local weighting but without attention pooling non-realizable. We further remove the global loss $\mathcal{L}_\text{global}$, i.e.\ set $\gamma=0$ and only use the local losses without local weighting, and observe a large drop in downstream performance. We assume that this is caused by missing contrast between samples.
Without the global loss, local weights can again not be computed, making a model with local weighting but without global loss non-realizable.

We also study the relevance of the local losses $\mathcal{L}_\text{local-image}$ and $\mathcal{L}_\text{local-report}$. Starting again from unmodified LoVT, we first adapt the local image loss $\mathcal{L}_\text{local-image}$ by redefining the positiveness score in a non-smooth way with $\ps \propto \mathds{1}_{[d_{\bm{x}}(k, l) \leq T]}$. The performance drops showing the relevance of the smoothness. 
When removing any of the local losses completely, i.e.\ either setting $\mu=0$ or $\nu=0$ and keeping only the global and one of the local losses ($\mathcal{L}_\text{local-image}$ or $\mathcal{L}_\text{local-report}$), the performance also drops compared to unmodified LoVT, showing that both local losses are required for optimal results. Note that removing $\mathcal{L}_\text{local-report}$ leads to a larger drop in downstream performance than removing only $\mathcal{L}_\text{local-image}$, indicating that $\mathcal{L}_\text{local-report}$ is more relevant for alignment.
When both local losses are fully removed by setting $\mu=0$ and $\nu=0$, such that only the global loss remains, the performance slightly drops compared to unmodified LoVT showing that the local losses are relevant components of the model. However, the drop in performance is smaller 
than when removing only one of the local losses, which indicates that the symmetry of the local losses is essential for them to work. 
If we further replace attention pooling by avg/max pooling, a large drop in performance is observed, which again highlights the importance of attention pooling. Note that without avg/max pooling the local losses provide more improvements than when using attention pooling.

We also study the differences to ConVIRT\cite{ConVIRT} and  (modified) CLIP\cite{CLIP}. Starting from ConVIRT, replacing their learning rate schedule (reducing on plateaus) by our cyclic cosine schedule (see \cref{sec:pre_training_details}) improves the results. Further replacing their avg/max pooling (to compute global representations) by attention pooling improves the 
results even further. This settings corresponds to (modified) CLIP.
In ConVIRT (and CLIP), only a single sentence per report is sampled when computing report representations. Replacing this sampling method by ours, where all sentences of a report are used to compute its representation, the results are improved if attention pooling is used. If no attention pooling is used, the performance degrades when using all sentences instead of a single randomly sampled one. 

In our ablation study we highlighted the importance of all components of LoVT. We also showed that some of our proposed changes can also be used to improve the ConVIRT or CLIP models.

\paragraph{Representation Distribution and Alignment Analysis}
\begin{figure}[t]
\centering
    \includegraphics[width=.5\linewidth]{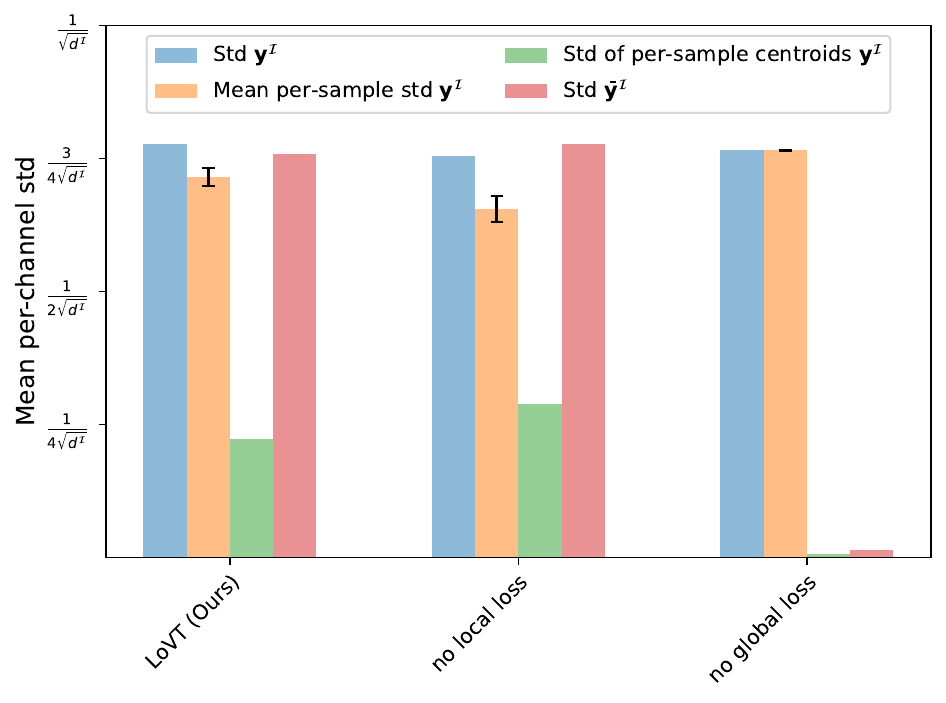}
        \caption{Standard deviation (std) of image ($\ygs$) and image region ($\ys$) representations. \textbf{Left}: LoVT (Ours) \textbf{Middle}: No local losses. \textbf{Right}: No global loss. For $\ys$ we additionally show the mean std per-sample, i.e.\ how different representations are within a sample, and the std of the per-sample centroids.
        The models were trained on 30\% of frontal MIMIC-CXR and then evaluated on the whole test set.}
        \label{fig:std}
\end{figure} 
\begin{figure}[t]    
    \centering
    \includegraphics[width=.5\linewidth]{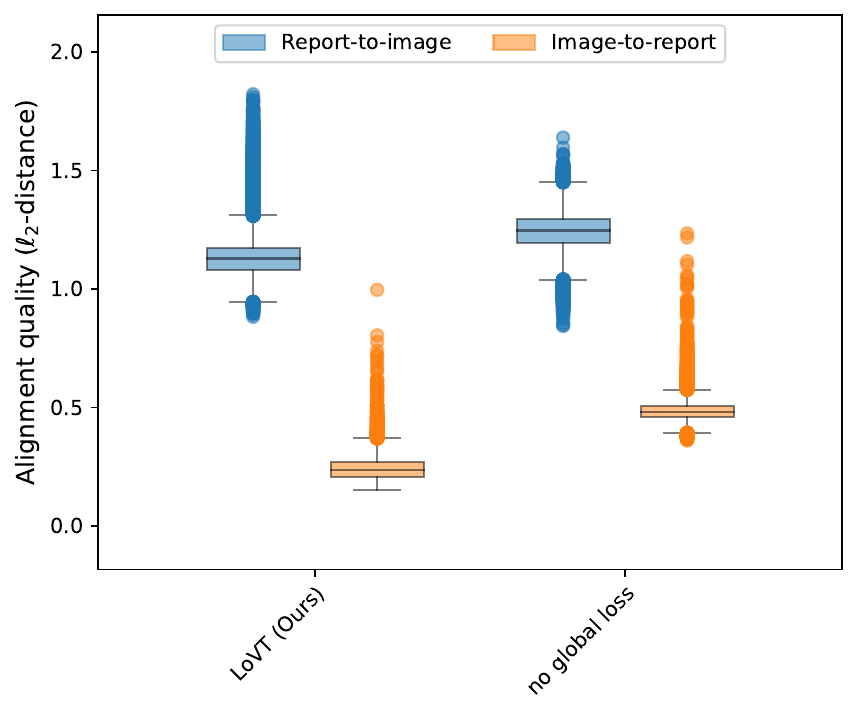}
        \caption{Alignment quality of local representations. 
        \textbf{Left}: LoVT (Ours) \textbf{Right}: No global loss.
        Measured by the $\ell_2$-distance of uni-modal with their related cross-modal representations. \textbf{Blue}: Report-to-image ($\zrs$) with image region ($\zs$) representations. \textbf{Orange}: Image-to-report ($\zsr$) with report sentence ($\zr$) representations. All representations are $\ell_2$-normalized before the distance is computed. The models were trained on 30\% of frontal MIMIC-CXR and then evaluated on the whole test set.}
        \label{fig:alignment}
\end{figure}
\begin{figure}[t]
    \centering
    \includegraphics[width=.5\linewidth]{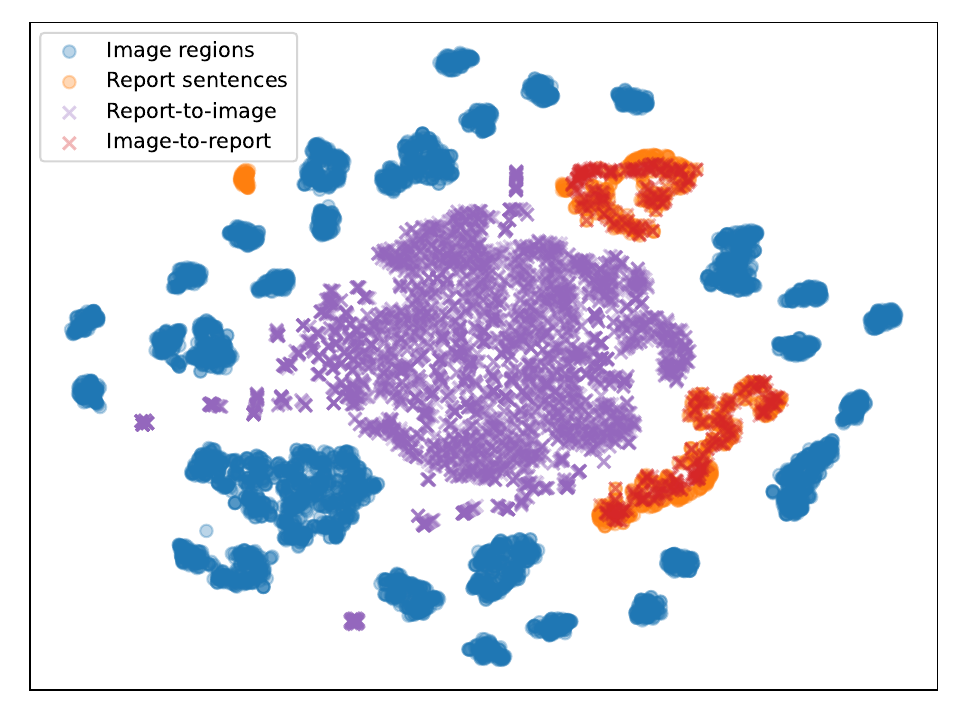}
        \caption{t-SNE\cite{tSNE} plot of projected local uni-modal representations (points) and the aligned cross-modal representations (crosses).
        \textbf{Blue}: Image regions ($\zs$). \textbf{Orange}: Report sentences ($\zr$).  \textbf{Purple}: Report-to-image ($\zrs$). \textbf{Red}: Image-to-report ($\zsr$).
        We trained our model on 30\% of frontal MIMIC-CXR and computed the representations on the first 100 samples of the test set.}
        \label{fig:tsne_zl}
\end{figure}
\begin{figure*}[t]
     \centering
     \subfloat{
         \includegraphics[width=0.3\linewidth]{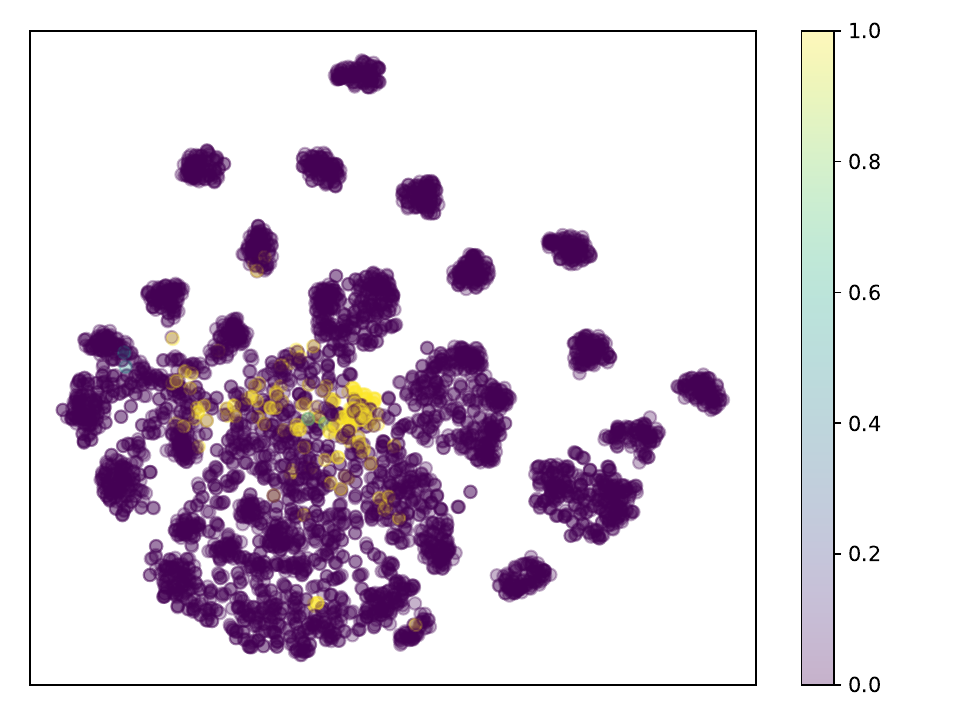}
     }
     \hfill
     \subfloat{
         \includegraphics[width=0.3\linewidth]{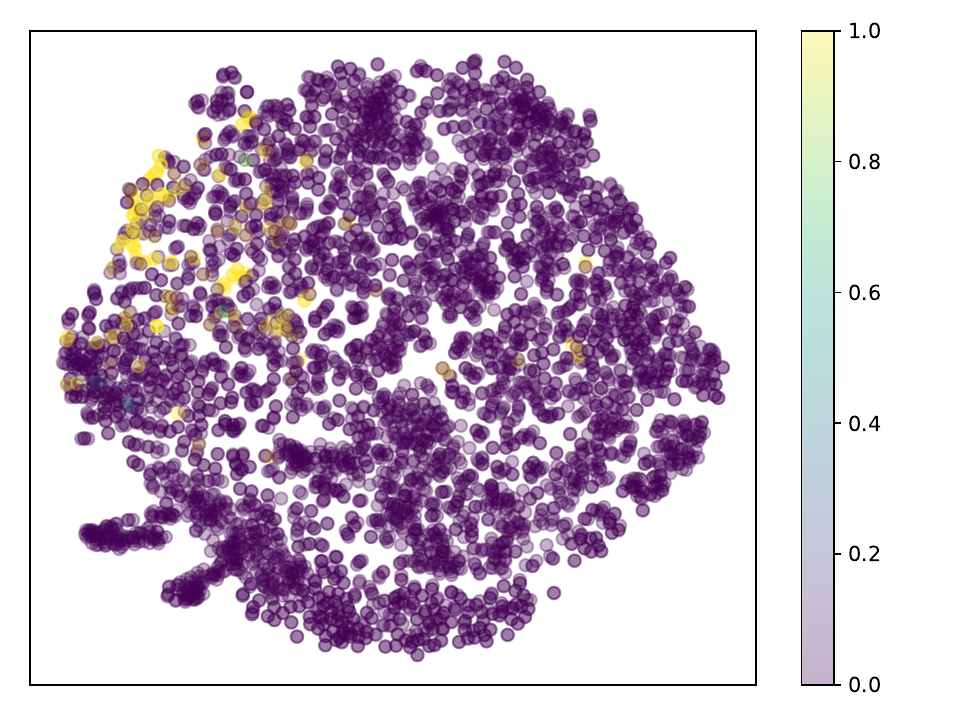}
     }
     \hfill
     \subfloat{
         \includegraphics[width=0.3\linewidth]{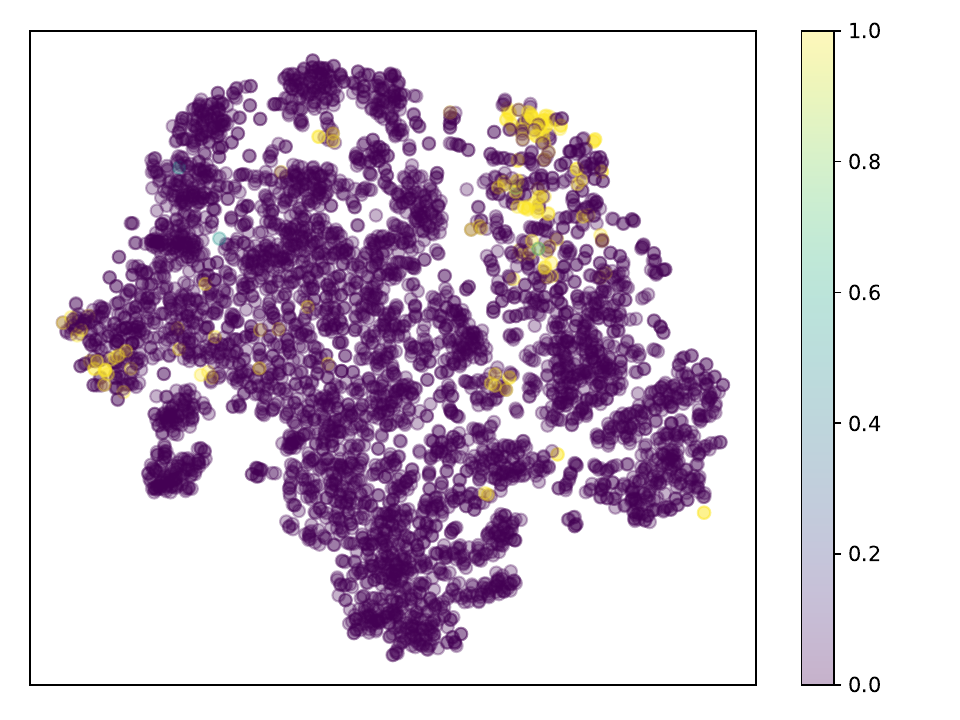}
     }
        \caption{t-SNE\cite{tSNE} plots of image region representations from samples of the RSNA pneumonia detection dataset. The color of each point indicates the overlap of the related region with a pneumonia opacity region. \textbf{Left}: LoVT (Ours). \textbf{Middle}: No local losses. \textbf{Right}: No local losses and no attention pooling. We trained all models on 30\% of frontal MIMIC-CXR and computed the representations on the first 100 samples of the RSNA test set.}
        \label{fig:rsna_l}
\end{figure*}

We analyze how representations are distributed and how well they are aligned. In \cref{fig:std} we show the standard deviation (std) of image $\ygs$ and image region $\ys$ representations of LoVT and variants of it without local losses or global loss. 
It can be observed that the (total) std of image region representations $\ys$ is similar in all three studied settings, indicating that the local and global losses have little influence on the overall variance of local representations.
We additionally analyze the mean per-sample std and std of per-sample centroids of $\ys$ to study how representations are distributed within a sample and between different samples, respectively. 
The per-sample std of $\ys$ is smallest when only using the global loss (no local losses) and largest when only using the local losses (no global loss). Vice versa, the centroids std is largest when only using the global loss (no local losses) and smallest when only using the local losses (no global loss). 
Therefore, the local losses encourage the representations to differ within each sample, i.e.\ ensure spatial sensitivity, while the global loss encourages them to differ between samples, i.e.\ prevents the collapse of per-image representations to a constant vector. 
The std of global image representations $\ygs$ behaves similarly to the centroids std of $\ys$, except that the local losses have only little influence on it. Note that the centroids std and std of global image representations almost completely vanish without the global loss, while there is still notable per-sample std present without the local losses. This highlights the importance of the global loss for preventing the collapse of representations.

In \cref{fig:alignment} we plot the alignment quality of local representations, i.e.\ the $\ell_2$-distance of report-to-image ($\zrs$) with their related image region representations ($\zs$) and of image-to-report ($\zsr$) with report sentence representations ($\zr$). We compare the representations learned by LoVT with (default) and without global loss. It can be observed that in both cases the image-to-report representations are better aligned than the report-to-image representations. This can be expected, as most of the information contained in the report is based on the image, making it easy to compute sentence representations from image region representations, while images contain additional details not described by the reports, making it harder to compute report-to-image representations. Both, report-to-image and image-to-report representations, are slightly better aligned when the global loss is used additionally to the local losses during training (as in the unmodified LoVT model). One can therefore deduce that the global loss supports the learning of local representations.

In \cref{fig:tsne_zl} we plot a t-SNE\cite{tSNE} projection of local representations learned by LoVT. Sentence representations are similarly distributed to image-to-report representations confirming, as already observed in the alignment analysis, that the model is able to align both distributions. Only one cluster of sentence representations is separated from image-to-report representations. We assume that these are sentences that do not describe features present in the image, e.g.\ describing features from lateral views or differences to previous studies of the patient. Image region representations and report-to-image representations are distributed differently, which again confirms that these could not be fully aligned. In the t-SNE\cite{tSNE} projection the image region representations are split into many clusters. We assume that this is a result of the negatives in the local image loss encouraging contrast between (spatially distant) region representations of each sample, such that our model behaves similarly to a clustering algorithm. %The high number of image region clusters may indicate an overclustering which may be caused by the non-optimal alignment of image region and report-to-image representations, although this is only an assumption.

To further study the effect of the local losses, we plot t-SNE\cite{tSNE} projections of image region representations from samples of the RSNA pneunomia detection dataset in \cref{fig:rsna_l}. We compare the representations learned using our unmodified LoVT model, our model without local losses, and our model without local losses and attention pooling.
For the unmodified LoVT model, image region clusters can again be observed while such clusters cannot be observed without the local losses. This confirms our assumption that these clusters are a result of the local losses.
It can also be observed that in the unmodified LoVT model the representations of pneumonia positive regions are distributed in a very confined area of space and are therefore probably easily separable from non-pneumonia regions. Without local losses the positive region representations are more spread over the space making them harder to separate. If attention pooling is not used as well, the positive region representations are distributed around multiple areas in space which may also hurt separability. Therefore, using the local losses and attention pooling improves separability of downstream representations which is confirmed by the results shown in \cref{tab:ablations}.

\paragraph{Local Weights}
\begin{figure}[t]
    \centering
    \includegraphics[width=.5\linewidth]{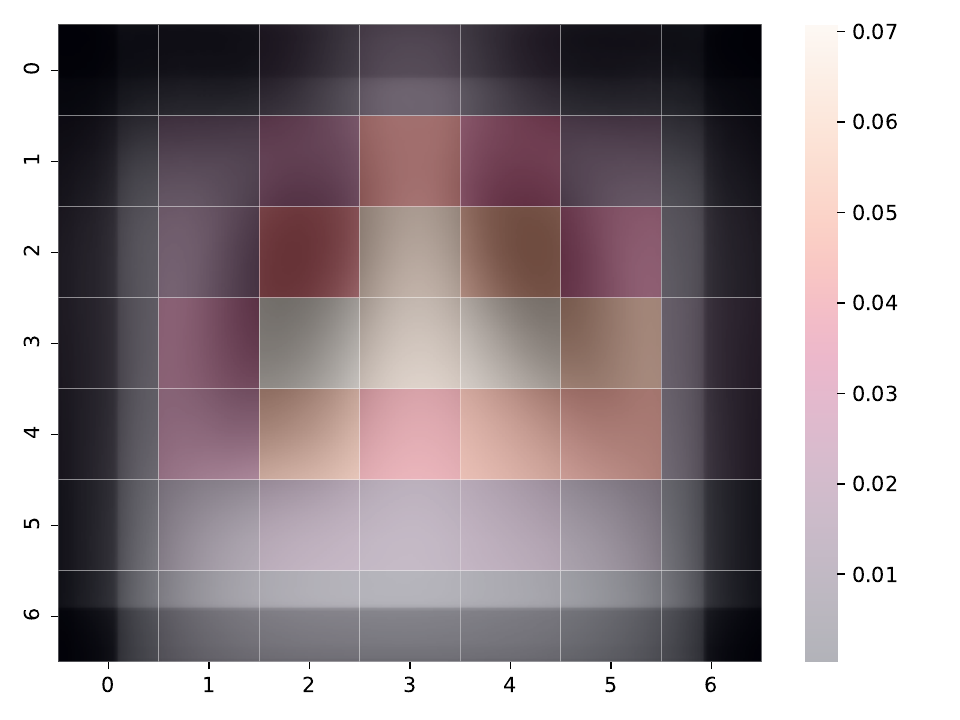}
        \caption{Local image region weights $\weights$ of different regions averaged over all samples and plotted on top of the mean image. The model was trained on 30\% of frontal MIMIC-CXR. Weights and mean image were computed on the whole test set.}
        \label{fig:l_weights}
\end{figure}

In order to understand how the weighting of image regions works, we study how the region weights $\weights$ are distributed. In \cref{fig:l_weights} we therefore plot the mean region weights on the mean image of the pre-training test set. The weights are distributed horizontally symmetrically around the center of the images and most focus is on the lungs and around the spine. This indicates that the weighting works as expected, as most pathologies in a frontal chest X-ray are typically observed at lungs or heart.

%\paragraph{Detailed Analysis of the NIH CXR Results}
%\begin{table*}[t]
%  \centering
%  \scriptsize
%  \setlength{\tabcolsep}{2pt}
%  \input{figures/results_table_nih}
%  \caption{Detailed results on the NIH CXR Linear task, showing differences between predicted classes.}
%  \label{tab:results_3}
%\end{table*}

\section{Discussion of the Limitations of LoVT}
\paragraph{Weak Supervision and Sensitivity to Hyperparameters}
As no supervision for the alignment of image regions and report sentences is available, we implicitly learn an alignment model in the latent representation space. We jointly learn this alignment model and the latent representations of image regions and report sentences, having only the global alignment information of image-report pairs as supervision. Therefore, we suspect that the model tends to converge to local optima, explaining its sensitivity to hyperparameters, especially to the learning rate. While using the cyclic-cosine learning rate schedule helps, our method is still hard to train and tune. 
We leave studying more explicit supervision, e.g.\ by including generative losses, to future work.

\paragraph{Limited Negatives for Local Alignment}
We only use local negatives from the same sample. By design, the number of local negatives is therefore very limited and many of these negatives may be very easy.
This may limit the model performance on downstream tasks\cite{SimCLR,MoCo}.
In preliminary experiments we also included negatives from other samples but could not observe a benefit. 
%We assume that is because for healthy regions or sentences these negatives are semantically related. 
We leave the study of losses with more negatives (e.g.\ based on the MoCo\cite{MoCo} approach) or without explicit negatives (e.g.\ based on the BYOL\cite{BYOL} approach) to future work.

\paragraph{Limited Alignment Model}
We decided to use single-head scaled dot product attention with linear projections as our alignment model. While this keeps the alignment model simple and computationally cheap, it also limits its capabilities. We leave studying more complex alignment models, like multi-head attention or (one or more) transformer\cite{transformer} layers, to future work.

\paragraph{Non-Adaptive Regions}
In LoVT the image region representations are computed for fixed regions, i.e.\ patches. Their boundaries are arbitrary and relevant features may therefore be spread across regions or multiple neighboring features may be in the same region, making it hard to learn region representations.
We leave studying other techniques for finding content-based regions and computing their representations to future work.

\section{Discussion of Negative Societal Impact}\label{app:societal_impacts}
In this section we discuss the possible negative societal impact of our work. We identified three primary aspects: i) usage of our method in medical applications, ii) data privacy issues, and iii) energy consumption.

\paragraph{Usage in Medical Applications}
As our method is targeted towards medical applications, potential issues in our method may lead to harm through misdiagnosis.
Most of the potential issues, including interpretability and reliability issues, are not specific to our method but apply to most deep learning methods in medicine and we therefore do not discuss them here. 
Still, we identified another potential issue: data bias learned during pre-training. While bias from data may be learned by most machine learning methods, in our case the bias might be learned from both, pre-training and fine-tuning data. During fine-tuning the pre-training dataset might not be available making it hard to identify such a data bias. 
As possible mitigation strategies the pre-training dataset (if available) should also be analyzed for data bias or the fine-tuned model should be tested for learned bias before using it in medical applications.
Note that this issue applies to most transfer learning approaches including other pre-training methods.

\paragraph{Data Privacy Issues}
Information learned from the pre-training dataset is contained in the pre-trained model weights, which may include information about individuals in the dataset. When distributing such models to make them available for others to fine-tune, this information is distributed as well. If the pre-training dataset is non-public but the pre-trained model is made publicly available, this may lead to data leakage and therefore a privacy breach. This is especially problematic if individuals can be re-identified. 
Therefore, pre-trained models should be distributed only under the same conditions as the pre-training dataset or other precaution measures, like privacy-preserving machine learning, should be taken to prevent data leakage. We thus decided not to release our pre-trained model weights publicly.
Note that this issue applies to most transfer learning approaches including other pre-training methods.

\paragraph{Energy Consumption and Environmental Impact}
Training of deep learning models consumes substantial amounts of energy and may therefore have environmental impacts. 
In our experiments, we observe that pre-training (including LoVT and the baselines) typically takes 0.5-2 days while downstream tasks typically only train for minutes to a few hours per run. While we did not study the exact energy consumption, we use the training times as an estimate and conclude that the energy consumption, and therefore the environmental impact, during pre-training is substantially higher than during finetuning. 
We observe that in our setting most studied pre-training methods, including LoVT, have similar runtimes (1-2 days, depending on the exact hyperparameters, for training on the full dataset) and only CheXpert pre-training is significantly faster (typically taking 0.5-1 days for pre-training). 
% chexpert 0.5d to 1d, LOVT: 1 to 1.5d, ConVIRT: 1 to 2d; simclr and BYOL: about 1.5d; pixel pro: about 2d
Besides training, also hyperparameter tuning needs to be considered, which may be required when applying LoVT to another pre-training dataset. 

While, as we observed, the high energy consumption of pre-training is an effect that is general to many methods, it should still be considered when deciding whether and how to use LoVT. An approach to reduce the energy consumption is to limit the hyperparameter tuning of LoVT on the pre-training dataset (e.g.\ only tune the learning rate) and use the defaults from our paper for most hyperparameters, although tuning other hyperparameters may improve downstream results. Instead, hyperparameter tuning could be more focused on the finetuning of the model. Additionally, pre-trained models should be made publicly available where this does not lead to privacy issues.

\section{Detailed Discussion of Related Work}
\subsection{Self-supervised Representation Learning} % https://arxiv.org/pdf/2006.08218.pdf
State-of-the-art methods for pre-training image models using self-supervised representation learning can be categorized into \emph{generative} and \emph{discriminative} approaches.
Generative models learn a distribution over the training images and a latent representation space. Typically, these approaches are autoencoding models\cite{denoiseAE,VAE,VQ-VAE,VQ-VAE2}, which learn to reconstruct the input image (or parts of it),
adversarial models\cite{GAN,BiGAN,ALI,GANLarge,BigBiGAN}, where data and representation are modeled jointly, autoregressive models\cite{PixelRNN,PixelCNN}, where image regions are conditioned on previous image regions, or flow-based models\cite{NICE,RealNVP,Glow}, which estimate high-dimensional densities from data. Generative models can recover the original data distribution without the need for assumptions on downstream tasks and are therefore well-suited for a wide range of applications, most notably for generative tasks\cite{SSL_summary}. However, they have some inherent problems, most notably, they model the distribution in the data space (e.g.\ in pixel-space) and therefore focus too much on low-level details (like pixels) instead of encouraging high-level abstractions that are typically required for discriminative downstream tasks like classification\cite{SSL_summary}.

Discriminative approaches are better suited for such tasks as they define discriminative objectives based on pretext tasks created from the unlabeled data.
Early discriminative approaches relied on heuristics when defining the pretext tasks\cite{pretext1,pretext2,pretext3,pretext4,pretext5}, limiting the generality of the resulting representations.
In recent years, contrastive approaches\cite{nonparam_disicrim,CPC,CPC2,DIM,misra2019selfsupervised,li2021prototypical,SimCLR,MoCo,BYOL,SimSiam,BarlowTwins,VICReg,W-MSE}, have become the state-of-the-art discriminative approaches for self-supervised representation learning. Contrastive methods act in the representation space and try to align representations of similar images (e.g.\ different views from the same image) while spreading representations of different images. Clustering approaches like DeepCluster\cite{DeepCluster} also belong to the contrastive approaches. 
Discriminative approaches are in general very lightweight and contrastive methods are currently state-of-the art for discriminative downstream tasks. However, they are not suited for generative tasks and many aspects, like the need for negative sampling, are not well-understood yet although being tackled by approaches like BYOL\cite{BYOL}, SimSiam\cite{SimSiam}, BarlowTwins\cite{BarlowTwins}, and VICReg\cite{VICReg}. Contrastive methods have also been successfully applied to medical imaging including image classification on chest X-rays\cite{gazda2021selfsupervised,sowrirajan2021mococxr,sriram2021covid19}.

Most contrastive learning approaches use instance-level contrast, i.e.\ represent each view of the image by a single vector. While the resulting representations are well-suited for global downstream tasks like image classification they lack properties like spatial sensitivity or smoothness required for more localized downstream tasks (like segmentation or detection)\cite{PixelPro}. Therefore, there is a number of recent approaches that use region-level contrast\cite{PixelPro,detco,DenseCL,chaitanya2020contrastive,pinheiro2020unsupervised,crosspixel_opticalflow}, i.e.\ they act on representations of image regions. These approaches are more suited for localized tasks and therefore typically outperform instance-level methods on such tasks.

\subsection{Multimodal Representation Learning}
While self-supervised representation learning on a single modality (e.g. images) already achieves great results, in some settings more modalities are available. Utilizing such additional modalities can improve the downstream results as additional information is available that can be utilized during representation learning. One form of such additional modalities is text, that often accompanies images in the form of captions or linked reports. Early works on combining image and text modalities did not focus on pre-training for downstream tasks but instead on learning aligned representations for cross-modal retrieval\cite{retrieval1,retrieval2}, on encoder-decoder tasks like image captioning\cite{caption1,caption2,caption3}, and on joint prediction tasks like visual question answering\cite{VisualQA} and visual grounding\cite{VisualGround1,VisualGround2}.
In recent years, several works utilized transformer\cite{transformer} models to compute joint representations of image content and text\cite{VideoBERT,VilBERT,VLBERT,LXMERT,UniVL,huang2020pixelbert}. They pre-trained their models using self-supervised tasks like multi-modal alignment prediction on paired image-text datasets and then finetuned them on multi-modal downstream tasks like image retrieval or visual question answering. 
While these methods can effectively pre-train joint image-text models, these models cannot be used for image-only downstream tasks. 
Recently, there is much focus on self-supervised representations learning methods that pre-train image models for downstream tasks by taking advantage of the companion text\cite{CLIP,ALIGN,ConVIRT,VirTex,ICMLM}. VirTex\cite{VirTex} and ICMLM\cite{ICMLM} use image captioning tasks (generative tasks), ConVIRT\cite{ConVIRT}, CLIP\cite{CLIP} and ALIGN\cite{ALIGN} use multiview contrastive learning\cite{AMDIM} (contrastive tasks). 
In\cite{wang2021selfsupervised} generative and contrastive losses are combined to train on mixed chest X-ray data, i.e. where only for some images paired reports is available.
LocTex\cite{liu2021loctex} does localized pre-training on natural images with companion text using a dot product based model to predict alignment of text and image regions. Unlike our method it uses supervision generated by mouse gazes instead of learning the alignment implicitly using a local contrastive loss.
Most related to our work is the recently published local-mi\cite{local_MI} that does contrastive learning on report sentences and image regions but aligns each sentence with its most related region instead of using an alignment model like our method. 
Also, it targets classification instead of localized tasks and does therefore neither encourage contrast between regions nor spatial smoothness.
%Due to the novelty of these approaches it is not yet clear whether generative or discriminative approaches are more promising in the multi-modal setting.
%The main difference to the transformer-based models is that these models require the text only for pre-training but not during downstream tasks such that they are suitable for image-only tasks. 

\section{Experiment Details}\label{app:experiment_details}
In all our experiments we use PyTorch\cite{pytorch} Version 1.10 (BSD-style license\footnote{\raggedright{\url{https://github.com/pytorch/pytorch/blob/master/LICENSE}}}) and train on a single NVIDIA Quadro RTX 8000.

\subsection{Pre-Training}
\label{sec:pre_training_details}
\paragraph{Pre-training Data and Pre-Processing}
Our method can be used with any dataset containing pairs of medical images and reports supposing that the reports contain multiple sentences and the sentences in the reports provide a semantically useful description of the contents in the image.
We use version 2 of MIMIC-CXR\cite{MIMIC-CXR-2,MIMIC-CXR,MIMIC-CXR-JPG,PhysioNet} as, to our best knowledge, it is the largest and most commonly used dataset of this kind conatining more than $200,000$ imaging studies, each with one or more frontal or lateral chest X-rays and one semi-structured free-text radiology report, written by a practicing radiologist during routine clinical care, describing radiological findings of the images.

We download the already pre-processed images from its JPG-version\footnote{\raggedright{\url{https://physionet.org/content/mimic-cxr-jpg/2.0.0/}} (PhysioNet Credentialed Health Data License)} and remove all except the frontal views, i.e.\ we only keep the \emph{antero-posterior (AP)} and \emph{postero-anterior (PA)} views.
We download the reports from MIMIC-CXR\footnote{\raggedright{\url{https://physionet.org/content/mimic-cxr/2.0.0/} (PhysioNet Credentialed Health Data License)}} and extract the text from the \emph{Findings} and \emph{Impression} sections. Reports containing none of these sections are removed.
For each report we concatenate the extracted text from both sections and remove reports where the extracted text contains less than three tokens (based on tokenizing it using Stanza\cite{stanza}).
We split the extracted text into sentences using Stanza\cite{stanza} again.
Finally, we remove all samples that contain no images or no report (after the previous steps) and then apply the training/validation/test splits provided by MIMIC-CXR-JPG\cite{MIMIC-CXR-JPG} such that we have 210228/1712/2867 training/validation/test samples (i.e.\ studies with one report and one or more images each), respectively.
%We only extracted the sentences of the findings and assessment sections of each report. As all downstream tasks only contain frontal views, we removed all lateral views.

\paragraph{Encoders and Model Details}
In the image encoder we use the ResNet50 implementation from Torchvision\footnote{\raggedright{\url{https://github.com/pytorch/vision} (BSD 3-Clause License)}} and initialize it with ImageNet\cite{ImageNet} weights\footnote{\label{resnet_url}\raggedright{\url{https://pytorch.org/hub/pytorch_vision_resnet/} (BSD 3-Clause License)}}.
In the report encoder we use the BERT\_base PyTorch implementation from Huggingface Transformers\cite{hf_transformers}\footnote{\raggedright{\url{https://github.com/huggingface/transformers} (Apache-2.0 License)}} and initialize it with weights from ClinicalBERT\cite{BioClinicalBERT}\footnote{\raggedright{\url{https://huggingface.co/emilyalsentzer/Bio_ClinicalBERT} (MIT License )}} that was trained on clinical notes.

We model the nonlinear transformations $\fs$, $\fr$, $\fgs$, and $\fgr$ as shallow neural networks without shared parameters, consisting of a (element-wise) linear layer with output size $2048$, batch norm, and ReLU followed by another linear layer with output size $512$.
This follows previous works\cite{SimCLR,BYOL,ConVIRT} except for the batch norm which we found beneficial.

\paragraph{Data Augmentation}
For image augmentations we first randomly sample one of the frontal chest X-rays of the sample (i.e.\ study) and then follow the augmentation scheme of ConVIRT\cite{ConVIRT}, i.e.\ random cropping (resized to $224 \times 224$), horizontal flipping, affine transformations, contrast and brightness jittering and Gaussian blur. We also tried removing geometric augmentations but found this setting to perform worse.
For text augmentations we concatenate all sentences of the Findings and Assessment sections of the report in the sample but randomly change the order by swapping pairs of sentences with a probability of $0.6$. We also tried randomly removing or duplicating sentences but did not find it to be beneficial.

\paragraph{Training Details and Cyclic Cosine Learning Rate Schedule}
For pre-training, we experimented with different learning rate schedules and found that a cyclic cosine learning rate schedule\cite{cosinelr} where the restarts also follow the cosine function and with a cycle length of two epochs (i.e.\ one decreasing and one increasing epoch) is beneficial. As both modalities have different properties (e.g.\ type of contained information) and the encoders have different architectures, they may converge at different speeds making it hard for both to adapt to each other. Therefore, we assume that the decreasing phase of the schedule allows the encoders to catch up and adapt to each other while the increasing phase allows them to learn faster and escape local optima.
We use the AdamW\cite{AdamW} optimizer with a batch size of $32$, with $16$ gradient accumulation steps, an initial learning rate of \SI{1e-4}, and weight decay \SI{1e-6} and train until the validation loss does not decrease for $10$ consecutive epochs.

\paragraph{Hyperparameter Tuning}
We tune the hyperparameters of LoVT using only the \emph{RSNA YOLOv3 Frozen 10\%} task (see \cref{sec:downstream_eval_details}).
For the hyperparameters of the global loss, i.e.\ $\tau$ and $\lambda$, and for the global representation dimension $\dgz$ we use the default values from ConVIRT\cite{ConVIRT}. In preliminary experiments, we tried different values for the local representation dimension $\dz$ but did not find that small changes to it have significant influence on the results, and therefore set $\dz = \dgz$ (i.e.\ 512). We determined the hyperparameters of the local losses, i.e.\ $\tau'$, $\beta$, and $T$, in preliminary experiments including grid searches and manual tuning. 
The loss weights $\gamma$, $\mu$, and $\nu$ were determined by running small grid searches in the following way: We first set $\gamma = \mu = \nu = 1$ and run a grid search to balance the local loss weights $\mu$ and $\nu$ while keeping $\gamma$ fixed, i.e.\ trying $(\mu = 0.5, \nu = 1.5)$, $(\mu = 1.0, \nu = 1.0)$, and $(\mu = 1.5, \nu = 0.5)$. After we found that $(\mu = 1.0, \nu = 1.0)$ performs best, we ran a grid search to balance local and global losses while keeping $\mu = \nu$, i.e.\ trying $(\gamma = 0.75, \mu = 1.0, \nu=1.0)$, $(\gamma = 1.0, \mu = 1.0, \nu=1.0)$, $(\gamma = 1.0, \mu = 0.75, \nu=0.75)$, and $(\gamma = 1.0, \mu = 0.25, \nu=0.25)$. We found that $(\gamma = 1.0, \mu = 0.75, \nu=0.75)$ performs best.

All hyperparameters except the learning rate are tuned using 30\% of the pre-training dataset and we tune the learning rate individually on 30\% and 100\% of the pre-training data. Note that we also slightly tune the learning rate when tuning other hyperparameters and in our ablation study.

\subsection{Baselines}
\paragraph{Random and ImageNet Init.}
For random initialization we do not pre-train the ResNet50 backbone but instead initialize it randomly following its default initialization.
For the ImageNet initialization we use the weights\footnoteref{resnet_url} provided by Torchvision.

\paragraph{CheXpert}
We train the ResNet50 backbone using multi-label binary classification on MIMI-CXR. We use five CheXpert\cite{chexpert} labels (Cardiomegaly, Edema, Consolidation, Atelectasis, and Pleural Effusion), which are included in the MIMIC-CXR-JPG\cite{MIMIC-CXR-JPG} dataset, and convert them to binary labels following the \emph{U-Ones} mapping\cite{chexpert} (i.e.\ mapping all uncertain labels to positive labels).
During CheXpert pre-training we use the full ResNet50 model including the average pooling and the fully connected (FC) layer but throw away the latter two for downstream tasks. 
All layers except the FC layer are initialized from ImageNet weights\footnoteref{resnet_url} and we randomly initialize the FC layer such that it has an output dimension of five (matching the number of classes). We use the sigmoid activation on the outputs, multi-label binary cross-entropy loss and the Adam\cite{Adam} optimizer and train with batch size 64 and weight decay \SI{1e-6} until the validation \emph{Area Under Receiver Operating Characteristic (AUROC)} does not increase for 10 consecutive epochs after which we select the checkpoint with the best validation AUROC. 
We tuned the initial learning rate and set it to \SI{3e-4} (\SI{1e-4} when trained on 30\% of the data). If the validation AUROC does not increase for three consecutive epochs we multiply the current learning rate by $0.5$.

\paragraph{SimCLR\cite{SimCLR}}
We use the PyTorch implementation available at \url{https://github.com/spijkervet/SimCLR} (MIT License) with the default image augmentations from the paper (with images resized to $224 \times 224$) except for color jittering where we do not adjust saturation and hue due to the monochrome nature of chest X-rays.
Following~\cite{ConVIRT} we set the output dimension to 128 and hidden size to 4096, use batch size 128, and weight decay \SI{1e-4}.
For training, we use the Adam\cite{Adam} optimizer and the cosine decay learning rate schedule\cite{cosinelr}, without restarts, over 100 epochs, with a single warm-up epoch.
We tuned the initial learning rate and set it to \SI{3e-4}.

\paragraph{BYOL\cite{BYOL}}
We use the PyTorch implementation available at \url{https://github.com/lucidrains/byol-pytorch} (MIT License)
with the default image augmentations from the paper (with images resized to $224 \times 224$) except for color jittering where we do not adjust saturation and hue due to the monochrome nature of chest X-rays.
We set the output dimension to 128 and hidden size to 4096, use decay rate $0.99$, batch size 64, and weight decay \SI{1e-4} .
For training, we use the Adam\cite{Adam} optimizer and the cosine decay learning rate schedule\cite{cosinelr}, without restarts, over 100 epochs, with a single warm-up epoch.
We tuned the initial learning rate and set it to \SI{1e-4} (\SI{3e-5} when trained on 30\% of the data).

\paragraph{PixelPro\cite{PixelPro}}
We use the PyTorch implementation available at \url{https://github.com/lucidrains/pixel-level-contrastive-learning} (MIT License) with the default image augmentations from the paper (with images resized to $224 \times 224$) except for color jittering where we do not adjust saturation and hue due to the monochrome nature of chest X-rays.

We set the output dimension to 512 and hidden size to 2048, use batch size 64, and weight decay \SI{1e-5}.
For training, we use the Adam\cite{Adam} optimizer and the cosine decay learning rate schedule\cite{cosinelr}, without restarts, over 100 epochs, with a single warm-up epoch.
We tuned the initial learning rate and set it to \SI{1e-3}.

\paragraph{ConVIRT\cite{ConVIRT}}
We use our own implementation of ConVIRT (as the general framework of ConVIRT is similar to LoVT) and
train until the validation loss does not decrease for 15 consecutive epochs after which we use the checkpoint with the lowest validation loss.
We tuned the learning rate and set it to \SI{1e-4} (\SI{1e-5} when trained on 30\% of the data). If the validation loss does not decrease for 12 consecutive epochs we multiply the current learning rate by $0.5$. We use the default values from the paper for all other hyperparameters.

\paragraph{CLIP\cite{CLIP}}
We use our own implementation of CLIP (as the general framework of CLIP is similar to LoVT). For better comparability with LoVT and the other baselines, we use ResNet50 and BERT\_base as encoders. Following the framework of CLIP we only encode single sentences and therefore randomly sample a sentence from the report (as in ConVIRT).
We use the AdamW\cite{AdamW} optimizer with a batch size of $32$ (the same as used in ConVIRT and LoVT), with $16$ gradient accumulation steps, and the cyclic cosine learning rate scheduler (as in LoVT) with an initial learning rate of \SI{1e-4}, and weight decay \SI{1e-6} and train until the validation loss does not decrease for $10$ consecutive epochs.

\paragraph{Batch Sizes of the Baselines}
Most contrastive learning methods are very sensitive to the used batch size, therefore the batch size is an important hyperparameter when comparing such methods. However, increasing the batch size also increases the GPU memory consumption and different methods have different memory requirements, such that using the same batch size for all methods does not allow for a fair comparison as in practice available GPU memory is typically limited.
We therefore decided to use three different batch sizes: The smallest batch size (32\footnote{We use this batch size as it was used in ConVIRT and as memory requirements are then kept below 24GB allowing training on widely used GPUs.}) is used for all text-supervised methods (i.e.\ ConVIRT, CLIP and our LoVT) as they require much memory due to their language model and they are also less sensitive to the batch size\cite{ConVIRT}. For image-only methods with a momentum encoder (i.e.\ BYOL and PixelPro) we use a larger batch size (64) and for SimCLR we further increase the batch size (128) as it does not have a momentum encoder and is very sensitive to the used batch size.

\subsection{Downstream Evaluation}
\label{sec:downstream_eval_details}
%\paragraph{Tuning and Evaluation Procedure}
%All baselines and our models have been tuned only on a single downstream task, \emph{RSNA YOLOv3 Frozen 10\%}, where a single fixed downstream learning rate was used (determined in prelimenary experiments) and the results of five runs have been averaged. Other downstream tasks have not been evaluated during tuning to make sure that models are not biased towards the downstream tasks.
%After tuning, each model was evaluated on all downstream tasks. For each task the learning rates were tuned individually per model (using single evaluation runs) before running five evaluations (all using the tuned learning rate). We report the average results of these five runs and their $95\%$-confidence interval (where each evaluation run is considered a sample).

\paragraph{Datasets}
\begin{itemize}
    \item \textbf{RSNA Pneumonia Detection}\cite{NIH_CXR,RSNA_2} (Licensed following the competition rules\footnote{\raggedright{\url{https://www.kaggle.com/c/rsna-pneumonia-detection-challenge/rules}}}): We download the dataset from its Kaggle page\footnote{\raggedright{\url{https://www.kaggle.com/c/rsna-pneumonia-detection-challenge/data}}} but use only their training set which we randomly split into our training, validation, and test set resulting in 16010/5337/5337 training/validation/test samples, respectively. For each sample we compute a segmentation mask (used in the \emph{Linear} evaluation) from all the ground truth detection boxes of that sample.
    \item \textbf{COVID Rural}\cite{COVID_Rural_1,COVID_Rural_2} (TCIA Data Usage Policy and CC BY 4.0 License): We download the dataset from its Github repository\footnote{\raggedright{\url{https://github.com/haimingt/opacity_segmentation_covid_chest_X_ray/tree/master/covid_rural_annot}}} and randomly split it into training, validation, and test set of sizes 133/44/44, respectively.
    \item \textbf{SIIM-ACR Pneumothorax Segmentation}\cite{siim_pneumo} (Licensed following the competition rules\footnote{\raggedright{\url{https://www.kaggle.com/c/siim-acr-pneumothorax-segmentation/rules}}}):
    We download the dataset from its Kaggle re-upload\footnote{\raggedright{\url{https://www.kaggle.com/seesee/siim-train-test/}}}, which is officially recommended on the original challenge website\footnote{\raggedright{\url{https://www.kaggle.com/c/siim-acr-pneumothorax-segmentation/overview/siim-cloud-healthcare-api-tutorial}}}, but use only their training set which we randomly split into our training, validation, and test set resulting in 7229/2409/2409 training/validation/test samples, respectively.
    \item \textbf{Object CXR}\cite{object_CXR} (CC BY-NC 4.0 License):
    We download the dataset from a re-upload\footnote{\raggedright{\url{https://academictorrents.com/details/fdc91f11d7010f7259a05403fc9d00079a09f5d5}}} as it is no longer available at its original source\footnote{\raggedright{\url{https://jfhealthcare.github.io/object-CXR/}}}. We randomly split their training set into our training and validation set and use their development set as our test set such that we have 6400/1600/1000 training/validation/test samples, respectively. For each sample we compute a segmentation mask (used in the \emph{Linear} evaluation) from all the ground truth detection boxes of that sample.
    \item \textbf{NIH CXR}\cite{NIH_CXR} (Licensed for public use with attribution\footnote{\raggedright{\url{https://nihcc.app.box.com/v/ChestXray-NIHCC/file/249502714403}}}):
    We download the ChestX-ray8 dataset provided by the NIH Clinical Center from its official website\footnote{\raggedright{\url{https://nihcc.app.box.com/v/ChestXray-NIHCC/}}} but use only the samples where bounding boxes are provided as ground truth.
    We randomly split these samples into our training, validation, and test set such that we have 588/196/196 training/validation/test samples, respectively.
\end{itemize}

\subsection{Evaluation Protocols}
In this section we describe the details of the evaluation protocols used in the evaluation framework\cite{eval}, including downstream model architectures and training details.
Note that we do not use image augmentations in any of the evaluation protocols but resize and pad the input images to size $224 \times 224$.

\paragraph{U-Net Finetune}
We do not use the original UNet\cite{UNet} architecture but instead build a UNet-like model\footnote{\raggedright{Our implementation is based on \url{https://github.com/kevinlu1211/pytorch-unet-resnet-50-encoder/blob/master/u_net_resnet_50_encoder.py} (MIT License)}} based on the pre-trained ResNet50 model. Therefore, we use the ResNet50 (except its avg pooling and FC layer) as the contracting path (left side) of our UNet-like model. The last feature map of ResNet50 has a size of $7 \times 7$ and dimension $2048$. Here we add two more convolutional blocks (each with a $3\times3$ convolution followed by batchnorm and ReLU) with output dimension $2048$ to the contracting path. For the expansive path (right side) we closely follow the architecture of the original UNet but use five instead of four upsampling blocks (each with $2\times2$ transposed convolution, concatenation, and two $3\times3$ convolutions each followed by batchnorm and ReLU) which have output dimensions $1024$, $512$, $256$, $128$, and $64$, respectively. For concatenation the ResNet50 blocks conv4, conv3, conv2, conv1, and the input image are used.
We add a single $1\times1$ convolution that predicts the positive class scores and use the binary Dice loss from the Segmentation Models Pytorch library\cite{seg_models_pytorch}.

For training, we use the Adam\cite{Adam} optimizer with weight decay \SI{1e-6}.
The learning rate is tuned individually for each model and task based on the best validation Dice\footnote{\label{dice_score}\raggedright{We use the micro-averaged Dice score based on this implementation: \url{https://torchmetrics.readthedocs.io/en/latest/references/modules.html\#f1} (Apache-2.0 License)}}. The learning rate is multiplied by $0.5$ if the validation Dice does not decrease for three consecutive epochs.
We use a warmup period in which we do not train the ResNet50 backbone but only the other, randomly initialized, layers with a learning rate of \SI{1e-3} after which we train the whole model (including the ResNet50).
On the COVID Rural dataset we use batch size eight, a warmup period of 20 iterations and do early stopping (based on validation Dice) after 20 epochs. On the SIIM-ACR Pneumothorax dataset we use batch size 64, a warmup period of 100 iterations and do early stopping after 10 epochs. Finally we report the test Dice of the epoch with the best validation Dice.

\paragraph{U-Net Frozen}
We use the same architecture and loss function as in the \emph{U-Net Finetune} protocol but freeze the pre-trained ResNet50 weights and never train them. Instead we only train the other layers using the same hyperparameters as in the \emph{U-Net Finetune} protocol (except for the warmup period which is not relevant in this setting).

\paragraph{Linear}
We use the frozen pre-trained ResNet50 (except for its avg pooling and FC layer) to compute $7\times7$ feature maps. A randomly initialized element-wise linear layer (i.e.\ a $1\times1$ convolution) is applied to these feature maps and the results are upsampled to the segmentation resolution using bilinear interpolation to predict the class scores. 
We then use the binary Dice loss from the Segmentation Models Pytorch library\cite{seg_models_pytorch}. For the NIH CXR dataset we train each class independently using the binary Dice loss. 

For detection tasks we first compute segmentation masks from the detection ground truth using the union of all target bounding boxes per sample and then interpret the task as a segmentation task. Note that for the Object CXR dataset we create bounding box masks only for box and ellipse detection targets but use polygon masks for polygon detection targets.

For training, where we only train the linear layer, we use the Adam\cite{Adam} optimizer with weight decay \SI{1e-6}.
The learning rate is tuned individually for each model and task based on the best validation Dice\footnoteref{dice_score}. The learning rate is multiplied by $0.5$ if the validation Dice does not decrease for three consecutive epochs.
On the \emph{COVID Rural Linear} and the \emph{RSNA Lin.\ Seg.\ 1\%} tasks we use batch size eight and do early stopping (based on validation Dice) after 20 epochs. On all other \emph{Linear} tasks we use batch size 64 and do early stopping after 10 epochs. Finally we report the test Dice of the epoch with the best validation Dice. Note that for the \emph{NIH CXR Linear} task we use the \emph{macro averaged Dice (Avg Dice)} as metric.

\paragraph{YOLOv3 Finetune}
We closely follow the architecture\footnote{\raggedright{Our implementation is based on \url{https://github.com/BobLiu20/YOLOv3_PyTorch}}} of the original YOLOv3\cite{YOLOv3} but use the pre-trained ResNet50 as its backbone (replacing the Darknet-53 backbone) while randomly initializing all other layers. The backbone features for the three prediction scales are extracted from the outputs of the conv3 (highest resolution), conv4, and conv5 (lowest resolution) blocks of ResNet50, respectively.
We use the default anchors presented in their paper but scale them according to our image input size of $224\times224$. 

For training, we use the losses and loss weights from the YOLOv3 paper and train with the Adam\cite{Adam} optimizer with weight decay \SI{1e-6}. 
The learning rate is tuned individually for each model and task based on the best validation \emph{mean Average Precision (mAP)}.
We compute\footnote{\raggedright{\url{https://github.com/bes-dev/mean_average_precision} (MIT License)}} the mAP score following the COCO\cite{COCO} mAP and with the following Intersection over Union (IoU) thresholds: $0.4, 0.45, 0.5, 0.55, 0.6, 0.65, 0.7, 0.75$. The learning rate is multiplied by $0.5$ if the validation mAP does not decrease for three consecutive epochs.
We use a warmup period of 100 iterations in which we do not train the ResNet50 backbone but only the other layers with a learning rate of \SI{1e-3} after which we train the whole model (including the ResNet50).
On the \emph{RSNA YOLOv3 Finetune 1\%} task we use batch size eight and do early stopping (based on validation mAP) after 20 epochs. On all other \emph{YOLOv3 Finetune} tasks we use batch size 64 and do early stopping after 10 epochs. Finally we report the test mAP of the epoch with the best validation mAP.

\paragraph{YOLOv3 Frozen}
We use the same architecture and loss functions as in the \emph{YOLOv3 Finetune} protocol but freeze the pre-trained ResNet50 weights and never train them. Instead we only train the other layers using the same hyperparameters as in the \emph{YOLOv3 Finetune} protocol (except for the warmup period which is not relevant in this setting).

\end{document}